\PassOptionsToPackage{table}{xcolor} %
\documentclass{article} %
\usepackage{iclr2027_conference,times}

\usepackage{amsmath,amsfonts,bm}

\def\eqref#1{equation~\ref{#1}}

\def\1{\bm{1}}

\DeclareMathAlphabet{\mathsfit}{\encodingdefault}{\sfdefault}{m}{sl}
\SetMathAlphabet{\mathsfit}{bold}{\encodingdefault}{\sfdefault}{bx}{n}

\usepackage{hyperref}
\usepackage{url}
\usepackage{enumitem} 
\usepackage{amsthm}
\newtheorem{definition}{Definition}

\usepackage{amsmath}
\usepackage{amssymb}
\usepackage{booktabs}
\usepackage{multirow}
\usepackage{graphicx}
\usepackage[table]{xcolor}
\usepackage{tikz}
\usetikzlibrary{arrows.meta, positioning, shapes.geometric, fit, backgrounds, calc}
\usepackage{pgfplots}
\pgfplotsset{compat=1.17}

\title{Know Your Context Bank: Repairing Inconsistency and Measuring Generalization in Context-Aware Text-to-SQL}

\author{Hailong Li, Mohammadreza Pourreza, Tal Neiman \\
\textbf{Reza Sherkat, Per Jacobsson, Yannis Papakonstantinou} \\
Google \\
\texttt{\{hailongli,pourreza,talneiman,rsherkat,pjacobsson,yannispap\}@google.com}}

\newcommand{\gain}[1]{\,{\scriptsize$(#1)$}}

\iclrfinalcopy %
\begin{document}

\maketitle
\lhead{}
\renewcommand{\headrulewidth}{0pt}
\begin{abstract}
Text-to-SQL systems increasingly rely on context banks: curated collections of example queries, metric definitions, and domain knowledge that tell a model how an organization writes its queries. In the enterprise, a bank grows over years as many analysts add entries, checked one at a time for correctness (does the SQL answer its own question?) and rarely for consistency with one another. Yet correctness is often loosely defined: a question seldom fixes how an entity is counted or which table defines it, so two entries can each pass review yet resolve the same decision in opposite ways. We call this condition \emph{context bank rot}. A second question follows: does a system that uses a bank generalize from it, or only reproduce it? Existing benchmarks cannot tell, as they do not control how far a test question sits from the context. We answer both questions by comparing queries decision by decision, with a fast scorer distilled from human-calibrated LLM judges. Turned inward, the comparison diagnoses and repairs a bank. Fine-tuning on a cleaned BIRD-train bank beats a same-size random subset at every epoch ($+2.6$ points on average), and $92.4\%$ of our removals had their SQL corrected by an independent expert audit, against $45.6\%$ of audited entries. In-context examples that follow one convention beat those following the opposite one by $13.3$ points, while a mix is barely better than none. Turned outward, the comparison generates test questions at a controlled distance from a bank, a pipeline we call CAT (Context-Aware Testing).
On CAT-BIRD, for retrieval and fine-tuning alike, the benefit of a bank falls step by step with distance and is largely gone once about three quarters of a question's logic is new, and on CAT-Spider the benefit of retrieval likewise shrinks with distance.
\end{abstract}

\section{Introduction}
\label{sec:intro}

Text-to-SQL systems in production rarely translate a question from the schema alone. They rely on a \textbf{context bank} (or simply \emph{bank}): a curated collection of vetted context, such as example queries, metric definitions, and domain knowledge, that shows the model how an organization defines its metrics and writes its queries \citep{shkapenyuk2025automatic}. We abstract every entry as a natural-language \emph{input} paired with the SQL that realizes it. The bank is either retrieved into the prompt at inference time \citep{arslan2024survey, gao2023text, pourreza2024din, pourreza2025chase} or compiled into training data for fine-tuning and reinforcement learning \citep{liu2026xiyan, pourreza2024dts, ma2026sql, yao2026arctic, pourreza2025reasoning}. Banks matter because even a question an analyst would consider complete does not fix its SQL: \emph{``What share of orders were refunded?''} can be reported as $0.42$ or as $42$, \emph{``How many customers bought a laptop?''} can count customer records or distinct customers, and \emph{``Which store sold the most?''} can return one row or every tied store. No analyst would ask for clarification; each organization settles on one way of writing such queries, and the bank records those conventions.

Anyone who maintains a context bank faces two challenges that cannot be assessed entry by entry: \emph{consistency}, whether the entries agree with one another, and \emph{generalization}, whether a system that uses the bank can carry its conventions to inputs the bank does not contain.

Consistency is invisible to the way banks are checked today. Entries accumulate over years from many analysts, onboarding documents, and query logs, and each is reviewed on its own terms: does this SQL answer this question? Benchmarks report correctness the same way \citep{yu2018spider, li2024can}, audits that found annotation bugs in them inspected entries in isolation \citep{wretblad2024bird, luoma2025snails, jin2026annotation}, and work on ambiguous questions resolves under-specification one question at a time \citep{bhaskar2023ambiqt, dong2025practiq}. Yet because a question seldom fixes how an entity is counted, which table defines it, or how edge cases are handled, two analysts can each write a defensible query and still make opposite choices for the same decision. In the BIRD \texttt{financial} database \citep{li2024can}, one entry defines ``accounts eligible for loans'' by account disposition (\texttt{disp.type = 'OWNER'}) while another joins the \texttt{loan} table; both execute cleanly and pass per-entry review, and a model trained or prompted with them receives two instructions for one decision. We call this condition \emph{context bank rot}. It belongs to the collection rather than to any entry, so no entry-level check can surface it. Section~\ref{sec:bank_health_index} shows that it is common in public banks, and Section~\ref{sec:hygiene_impact} that it costs accuracy.

We make consistency checkable by comparing entries with one another rather than each entry with its question. We decompose every entry's SQL into \emph{semantic units}, the individual decisions it makes (an aggregation, a filter, a join, an output shape), each tied to the span of the question that licenses it, and ask of any two entries, decision by decision, whether they agree, conflict, or address different things. A fast local scorer, distilled from human-calibrated LLM judges, computes this \emph{align--conflict--novel} profile for every pair. Aggregated over a bank, the profiles yield a \textbf{Bank Health Index} and point to the entries that need adjudication, and a two-track protocol repairs the bank: one track audits each entry against its own question, the other adjudicates conflicting pairs against an explicit rulebook of conventions.

Generalization is just as hard to see. Benchmarks do not control how far a test question sits from the context a system was given, so a system that has memorized its bank and one that has learned from it can look the same. Perturbation suites \citep{gan2021towards, deng2021structure, chang2023dr} and compositional splits \citep{keysers2020measuring, shaw2021compositional, gan2022measuring} do not relate each test question to a specific bank (Appendix~\ref{sec:related_work}). Our pairwise profile fills this gap: its alignment component is a graded distance between two queries, so the tool that finds conflicts inside a bank can also place a new question at a chosen distance from it. We turn this into \textbf{CAT} (\textbf{C}ontext-\textbf{A}ware \textbf{T}esting), a pipeline that expands a clean bank with test questions generated for a target distance, kept only if the scorer measures that distance, and selected so that no-context difficulty matches across distances. Accuracy as a function of distance then shows how far a system's use of the bank reaches.

Our contributions are as follows.
\begin{enumerate}[leftmargin=*, itemsep=1pt, topsep=1pt]
  \item \textbf{We identify cross-entry consistency as a data quality problem distinct from per-entry correctness.} Entries that each pass review can still encode contradictory conventions, a condition we call context bank rot and measure on public banks (Section~\ref{sec:bank_health_index}).
  \item \textbf{We propose a method to measure and repair inconsistency}: semantic units, the align--conflict--novel profile, a distilled pairwise scorer, the Bank Health Index, and a two-track cleaning protocol (Sections~\ref{sec:semantic_foundation}--\ref{sec:bank_health_index}). Cleaning BIRD-train improves fine-tuning of Qwen2.5-7B by $2.6$ points on average over a same-size random subset, and with every entry defensible, in-context examples that follow one convention beat the opposite convention by $13.3$ points, while a mix forfeits most of the gain (Section~\ref{sec:hygiene_impact}).
  \item \textbf{We turn the same measurement into CAT, a pipeline that builds benchmarks for how far context-aware systems generalize.} On CAT-BIRD ($1{,}097$ questions at scorer-verified distances from a $1{,}501$-entry bank), the benefit of the bank decays step by step with distance for retrieval and fine-tuning alike and is largely gone once three quarters of a question's logic is new; fine-tuning hard enough to reproduce the bank can cost accuracy on databases the bank does not cover. On CAT-Spider ($953$ questions from a $1{,}833$-entry Spider bank), the benefit of retrieval likewise shrinks with distance (Section~\ref{sec:cat_benchmark}).
\end{enumerate}

\section{Measuring How Two Entries Relate}
\label{sec:semantic_foundation}

A \textbf{context bank} $\mathcal{B} = \{(Q_i, S_i)\}_{i=1}^{N}$ pairs each natural language question $Q_i$ (with any domain notes or value mappings that accompany it) with a vetted SQL query $S_i$ over a schema $\mathcal{D}$; a context-aware system retrieves a subset of $\mathcal{B}$ into its prompt or fine-tunes on all of it. Because neither question nor schema settles every decision in a query, entries can resolve the same decision differently, and a new question can reuse some of an entry's decisions while adding others; this section builds a tool that measures both.

\subsection{Semantic Units: The Decisions a Query Makes}
The BIRD entry \emph{``How many schools in Amador which the Low Grade is 9 and the High Grade is 12?''} (SQL in Figure~\ref{fig:acn_comparison}a) makes five separable decisions: count schools, filter on county, filter on low grade, filter on high grade, and join \texttt{frpm} to \texttt{schools}. Four are asked for by a phrase in the question; the join is not mentioned but is forced by the schema. We call each such decision a \emph{semantic unit}.

\begin{definition}[Semantic Unit]
A semantic unit is a tuple $u = (\tau, \kappa, \sigma, \nu, \gamma)$ of a \textbf{type} $\tau$ (\texttt{aggregation}, \texttt{predicate}, \texttt{join\_edge}, \texttt{group\_by}, \texttt{order\_by}, \texttt{distinct}, or \texttt{output\_shape}); a \textbf{key} $\kappa$ naming the schema elements involved (e.g., \texttt{pred:schools.county:eq}), used to match units across entries; the \textbf{SQL fragment} $\sigma$ that realizes the decision (e.g., \texttt{schools.County = 'Amador'}); the \textbf{licensing span} $\nu \subseteq Q$, the verbatim phrase in the question that asks for it (e.g., \emph{``in Amador''}); and a \textbf{grounding} $\gamma$, \texttt{verbalized} if the decision is licensed by text and \texttt{structural} if only by the schema, as with a foreign-key join.
\end{definition}

\texttt{sqlglot} reads each fragment $\sigma$ and key $\kappa$ off the syntax tree; an LLM then assigns each fragment its span $\nu$ and grounding $\gamma$, and every span must be an exact substring of $Q$, so the LLM cannot invent a justification the question does not contain. Units also give a per-entry test, \emph{semantic licensing}: every fragment must be licensed by the question or the schema, and every requirement in the question must be realized. A violation is a defect of a single entry, which Track~A (Section~\ref{sec:two_track}) audits for.

\subsection{Comparing Two Entries: The Align--Conflict--Novel (ACN) Profile}
Given two entries $e_A$ and $e_B$, we pair up units that play the same \emph{role} in their queries (both filter on county, both count schools), matching as many as possible, and place each role in one of three buckets. A role is \textbf{aligned} when both units realize it with the same SQL fragment; if the fragments differ only in a literal that each question names for itself (\emph{``in Amador''} vs.\ \emph{``in Alameda''}), the entries simply asked different things, and the role counts as novel rather than as a conflict. A role is \textbf{conflicting} when the two units realize it with different SQL that their questions do not explain, as when both entries ask for ``accounts eligible for loans'' but one filters on \texttt{disp.type} and the other joins \texttt{loan}. A role is \textbf{novel} when it appears in only one entry.
Let $\rho$ be the number of distinct roles across the pair, counting each matched pair once and each unmatched unit once. With $w_{\text{align}}(r), w_{\text{conf}}(r) \in [0,1]$ the aligned and conflicting credit of a matched role $r$, the \textbf{ACN profile} of the pair is
\begin{equation}
  \alpha = \frac{1}{\rho}\sum_{r \text{ matched}} w_{\text{align}}(r), \qquad
  \beta  = \frac{1}{\rho}\sum_{r \text{ matched}} w_{\text{conf}}(r), \qquad
  \omega = 1 - \alpha - \beta,
  \label{eq:acn}
\end{equation}
so the three components sum to one and unmatched roles contribute only to $\omega$. A value $\beta > 0$ signals disagreement on a shared decision, and $\alpha$ measures how much of one entry is reproduced in the other, which Section~\ref{sec:cat_benchmark} uses as a distance. In Figure~\ref{fig:acn_comparison}a, four of five roles are identical and the county differs by a licensed value, so the pair differs without disagreeing, $\mathbf{p} = (0.80, 0.00, 0.20)$; in Figure~\ref{fig:acn_comparison}b, the entries share one role and realize it incompatibly, giving $\beta = 1/3$. A near-duplicate pair has high $\alpha$, while the typical pair in a healthy bank shares no roles at all (median $\omega \approx 0.99$).

\begin{figure*}[t]
\centering
\resizebox{\textwidth}{!}{%
\footnotesize
\setlength{\tabcolsep}{4pt}
\renewcommand{\arraystretch}{1.02}
\begin{tabular}{p{0.49\textwidth} p{0.49\textwidth}}
\toprule
\textbf{(a) Licensed Value Change} $\mathbf{p} = (\mathbf{0.80}, 0.00, 0.20)$ & \textbf{(b) Convention Conflict} $\mathbf{p} = (0.00, \mathbf{0.33}, 0.67)$ \\
\midrule
$e_A/e_B$: ``\emph{Schools \textbf{in Amador} / \textbf{in Alameda} with Low Grade 9, High Grade 12?}'' \newline
\hspace*{1mm}\texttt{\scriptsize SELECT COUNT(T1.`School Name`) ... WHERE T2.County = 'Amador'|'Alameda'} &
$e_A$: ``\emph{Accounts in Tabor \textbf{eligible for loans} (type='OWNER').}'' \newline
\hspace*{1mm}\texttt{\scriptsize ... JOIN disp T3 ON T1.account\_id = T3.account\_id WHERE T3.type = 'OWNER'} \newline
$e_B$: ``\emph{How many accounts in Prague are \textbf{eligible for loans}?}'' \newline
\hspace*{1mm}\texttt{\scriptsize ... INNER JOIN loan T3 ON T1.account\_id = T3.account\_id} \\
\textbf{Reading:} 4 of 5 roles identical; the county differs, but each question names its own county, so $\beta = 0$. &
\textbf{Reading:} One shared role: $e_A$ follows its domain note (\texttt{disp}), $e_B$ joins \texttt{loan}. Both pass per-entry review; only the comparison reveals that the bank defines one concept two ways. \\
\bottomrule
\end{tabular}%
}
\caption{Two entry pairs with their ACN profiles $\mathbf{p} = (\alpha, \beta, \omega)$. A value that each question names for itself is a difference but not a conflict (a); a shared role realized two incompatible ways is a conflict (b). Both pairs are real and taken from BIRD.}
\label{fig:acn_comparison}
\end{figure*}

\section{Distilling the Judge System into a Pairwise Scorer}
\label{sec:scorer_distillation}

An LLM can compute the ACN profile of a pair directly, but auditing a bank means scoring every same-database pair, more than $10^5$ for BIRD-dev alone, which is too slow and costly for a frontier model at every audit. We therefore use frontier LLMs only to label training pairs, and distill them into a local scorer.

\subsection{Labeling Pairs with an Ensembled Judge System}
\label{sec:judge_labels}
Reference profiles come from two independent LLM judges, Claude-Sonnet-5 and Gemini-3.5-Flash. Each is given the two entries and the schema they touch, decomposes each query into semantic units, matches units across the pair, and returns a profile $\mathbf{p} = (\alpha, \beta, \omega)$ with a short rationale for each matched role. When the judges disagree we want neither guess in the training data, so we keep a pair only when they agree closely:
\begin{equation}
  \text{Label}(\mathbf{p}_1, \mathbf{p}_2) =
  \begin{cases}
   \text{abstain}, & \text{if } |\alpha_1 - \alpha_2| > 0.20 \text{ or } \big(\max_j \beta_j \ge 0.15 \text{ and } \min_j \beta_j \le 0.05\big),\\[2pt]
   \tfrac{1}{2}(\mathbf{p}_1 + \mathbf{p}_2), & \text{otherwise.}
  \end{cases}
  \label{eq:consensus_policy}
\end{equation}
The second condition discards pairs where one judge sees a conflict and the other does not, exactly where a single-judge label is least reliable; abstained pairs are dropped (Appendix~\ref{app:calibration} shows one pair of each outcome). The judges' instructions and this rule were calibrated against a human expert over fourteen rounds, the last ten blind: the expert adjudicated every disagreement between anonymized, shuffled judge labels, and each round's adjudications changed the instructions or the rule (Appendix~\ref{app:judge_calibration}). By the last rounds the expert found almost nothing to correct.

\subsection{Training Data, Model, and Accuracy}
We build training pairs from verified BIRD-train entries, holding out BIRD-dev entirely: \emph{real pairs} ($63\%$), two entries from the same database, from near-identical to unrelated, labeled by the judges; \emph{constructed variants} ($33\%$), an entry whose SQL or question is edited in a controlled way (swap a value, drop or add a predicate, change \texttt{>} to \texttt{>=}, bridge two entries), with a profile known by construction; and \emph{rephrases} ($4\%$), the same SQL under a reworded question, with profile $(1, 0, 0)$. Constructed pairs keep the scorer from learning from the judges alone. Adding $4{,}403$ BIRD-train pairs converted to PostgreSQL, $730$ LiveSQLBench pairs, and $238$ constructed Spider-train pairs gives $20{,}568$ labeled pairs over $208$ databases, each used in both orders. The scorer is Qwen3.6-27B with a two-output head, fine-tuned with LoRA on the schema fragment followed by the two entries (over-long inputs lose only schema description). The head outputs $\hat{\alpha}$ and $\hat{\beta}$ through sigmoids trained with binary cross-entropy against the soft labels, and $\hat{\omega} = \max(0, 1 - \hat{\alpha} - \hat{\beta})$ (Appendix~\ref{app:scorer}).

On $2{,}879$ held-out pairs from the $11$ BIRD-dev databases, built and labeled the same way, $\hat{\alpha}$ has a mean absolute error of $0.020$ and a Spearman correlation of $0.970$ with the label, and $\hat{\beta}$ separates conflicting from non-conflicting pairs with AUC $0.982$. Real pairs are harder ($1{,}690$ pairs; Spearman $0.855$, AUC $0.872$; Appendix~\ref{app:scorer}). We use $\hat{\alpha}$ as a graded distance (Section~\ref{sec:cat_benchmark}). $\hat{\beta}$ ranks pairs reliably but is not a calibrated probability, as its values are compressed relative to the judges' (Appendix~\ref{app:scorer}), so it only decides which pairs to look at; the judge system decides what to do with them.

\section{Diagnosing and Cleaning a Context Bank}
\label{sec:bank_health_index}

\subsection{From Pairwise Profiles to a Bank Health Index}
Running the scorer over every pair of entries from the same database gives each pair an alignment score $\hat{\alpha}$ and a conflict score $\hat{\beta}$. Most pairs concern unrelated questions and say nothing about consistency, so we look only at \emph{overlapping} pairs, those that share at least a tenth of their decisions ($1-\hat{\omega} \ge 0.1$). Among them, the \textbf{conflict rate} $c$ is the share flagged as conflicting ($\hat{\beta} \ge 0.05$), and the \textbf{duplicate rate} $d$ is the share in which one entry reproduces at least half of the other's decisions ($\hat{\alpha} \ge 0.50$). The \textbf{Bank Health Index} is
\begin{equation}
  \text{BHI}(\mathcal{B}) = 100 \cdot (1-c)^2 \cdot (1-d).
  \label{eq:bhi}
\end{equation}
A bank with no flagged conflicts and no duplicates scores $100$. Conflicts weigh about twice as much as duplicates, since $(1-c)^2 \approx 1-2c$, because a duplicate wastes a retrieval slot or a training example but teaches nothing wrong, whereas a conflicting pair teaches two answers to one decision. Low overall alignment is deliberately not penalized: a bank covering many questions \emph{should} have most pairs unrelated. Because $c$ and $d$ are shares of pairs, removing entries at random leaves them unchanged in expectation, so banks of different sizes can be compared directly; a random half of BIRD-train has $c = 0.052$ against $0.053$ for the full pool. Within a bank, an entry that conflicts with many of its overlapping peers (a \emph{hub}) is the usual signature of a wrong entry or of a minority convention.

\begin{table}[t]
\centering
\caption{Bank Health Index of four public context banks before cleaning (rates are shares of overlapping pairs). Spider-train: medium and harder questions in its $42$ largest databases. LiveSQLBench ($15$ entries per database) is pooled. BIRD-dev is held out from scorer training; the other banks contributed training pairs (the $6$ held-out LiveSQLBench databases alone give $c = 0.235$, BHI $58.6$).}
\label{tab:bhi_results}
\footnotesize
\setlength{\tabcolsep}{3pt}
\resizebox{\textwidth}{!}{%
\begin{tabular}{llrrccc}
\toprule
\textbf{Bank} & \textbf{Dialect} & \textbf{Entries} & \textbf{Overlapping pairs} & \textbf{Conflict $c$} & \textbf{Duplicate $d$} & \textbf{BHI} \\
\midrule
BIRD-train (69 DBs) & SQLite & $6{,}980$ & $81{,}398$ & $0.053$ & $0.033$ & $86.8$ \\
BIRD-dev (11 DBs) & SQLite & $1{,}534$ & $20{,}245$ & $0.051$ & $0.016$ & $88.6$ \\
Spider-train, subset (42 DBs) & SQLite & $3{,}001$ & $29{,}210$ & $0.008$ & $0.111$ & $87.4$ \\
LiveSQLBench (18 DBs) & PostgreSQL & $270$ & $203$ & $0.187$ & $0.000$ & $66.1$ \\
\bottomrule
\end{tabular}%
}
\end{table}

\paragraph{What the index finds.}
Table~\ref{tab:bhi_results} scores four public banks. BIRD-train and BIRD-dev have similar conflict rates ($5.3\%$ and $5.1\%$ of overlapping pairs), but the rate varies widely across databases, from $1.0\%$ in \texttt{student\_club} to $22.2\%$ in \texttt{financial}. LiveSQLBench, whose entries are long analytical queries, reaches $18.7\%$ with no near-duplicates. Spider-train is the most consistent bank ($0.8\%$ conflicts); its index is held down by duplicates instead, since $11.1\%$ of its overlapping pairs are rewordings with the same SQL. Elsewhere duplicates are rare (at most $3.3\%$), so conflicts dominate the index, and our cleaning experiments use BIRD, where conflicts are common. A flag marks tension, not a confirmed contradiction, and the judge system finds causes that run from ``one entry is wrong'' to ``both are defensible.'' At one end are \emph{annotation bugs}, such as an entry asking about ``merged Alameda'' whose SQL filters \texttt{County = 'Lake'} and so conflicts with seven correct peers. At the other end are \emph{convention splits}, such as the two definitions of ``eligible for loans'' in Figure~\ref{fig:acn_comparison}b or different subquery structures for the same computation, and \emph{schema ambiguity}, where two columns (a city name and a city code) both plausibly play the same role. The two ends call for different remedies, which is why our cleaning protocol has two tracks.

\subsection{The Two-Track Cleaning Protocol}
\label{sec:two_track}
A flag says where to look, not what to remove, for two reasons. First, most flagged pairs are not contradictions: of all $252$ pairs in \texttt{california\_schools} with $\hat{\beta} \ge 0.03$, only $4$ posed the same request; the rest are different questions that share a structural template (Appendix~\ref{app:calibration}). Second, many defective entries raise no flag at all: an entry that is wrong about a question no other entry asks has no neighbor to disagree with. The protocol therefore uses the scorer only to choose which pairs deserve a closer look, lets LLM referees decide what to remove, and removes nothing without a concrete, recorded reason. An entry is removed if either track removes it.

\paragraph{Track A: audit each entry on its own.} Every entry, flagged or not, is audited by the judge system (§\ref{sec:judge_labels}), which sees the entry's question, its SQL, and the result of running that SQL, and removes the entry only if the ensemble agrees that the SQL does not do what its own question asks; stylistic choices are never flagged (checklist in Appendix~\ref{app:wash}). On BIRD-dev, $331$ entries are called a bug by at least one judge, but only $97$ are removed.

\paragraph{Track B: settle disagreeing pairs with a rulebook.} Every same-database pair with $\hat{\beta} \ge 0.03$ (on BIRD-train, within each entry's $20$ nearest neighbors) goes to the judge system, and a pair with a confirmed conflicting unit goes to an adjudicator that stops at the first check that applies: if the questions ask for different things, both entries are kept; if one side is wrong for its own question, it is removed; if a rule prefers one side, the other is removed and the rule recorded; otherwise both are kept for a person to decide. The rulebook lists ten conventions, written by us, for choosing between two defensible realizations (Table~\ref{tab:rulebook}); its rules name no table or database, so one rulebook serves every bank, and a curator who disagrees with a rule can edit it and rerun this step without rescoring.

\paragraph{What gets removed.} Most of Track B's work is clearing false alarms: on BIRD-dev, three quarters of its pairs end with both entries kept (Table~\ref{tab:trackb_outcomes}). The two tracks remove $157$ of $1{,}534$ BIRD-dev entries ($10.2\%$) and $1{,}332$ of $6{,}980$ BIRD-train entries ($18.9\%$), most by Track A, each recorded in a per-entry ledger with its track, rule, and the referees' reasons. As an external check, an independent expert audit of BIRD-train (BIRD-Platinum; \citealp{zhu2026revisql}) corrected the SQL of $45.6\%$ of the $1{,}821$ questions it shares with our pool, but of $92.4\%$ ($317$ of $343$) of our removals among them, so the removals are conservative but precise (Appendix~\ref{app:wash}).

\section{Impact of Context Bank Hygiene: Training and Inference}
\label{sec:hygiene_impact}

We test cleaning in both ways a bank is used. The training experiment removes defects of every kind, so it measures what cleaning buys as a whole. The inference experiment uses only entries that are each defensible, so any effect it finds comes from consistency alone.

\subsection{Training on a Cleaned Context Bank}
\paragraph{Setup.}
We fine-tune Qwen2.5-7B with LoRA (rank $128$, learning rate $2\times 10^{-5}$, cosine schedule) for four epochs on three versions of the $6{,}980$-entry BIRD-train pool and compare them on BIRD-dev ($1{,}534$ questions) by execution accuracy. The \textbf{cleaned} set is the $5{,}648$ entries that survive the protocol (§\ref{sec:two_track}). The \textbf{random} set, the natural control, draws $5{,}648$ entries at random, stratified by database and SQL length, so it keeps defective entries at their natural rate. The \textbf{bad-kept} set keeps every entry the protocol would remove and deletes an equal number of clean ones instead, as a dose-response check (Appendix~\ref{app:wash}).

\begin{table}[t]
\centering
\caption{Execution accuracy (\%) on BIRD-dev ($n = 1{,}534$) of Qwen2.5-7B fine-tuned on each training set; BHI scores the training set, and the last two columns split Epoch 3 by BIRD difficulty.}
\label{tab:sft_wash_ablation}
\resizebox{\textwidth}{!}{%
\footnotesize
\setlength{\tabcolsep}{6pt}
\begin{tabular}{lcccccccc}
\toprule
\textbf{Training set} & \textbf{BHI} & \textbf{Base} & \textbf{Ep.\ 1} & \textbf{Ep.\ 2} & \textbf{Ep.\ 3} & \textbf{Ep.\ 4} & \textbf{Ep.\ 3, moderate} & \textbf{Ep.\ 3, challenging} \\
\midrule
Bad-kept ($n=5{,}650$) & $85.9$ & $46.5$ & $49.7$ & $51.4$ & $50.3$ & $50.2$ & $40.9$ & $22.9$ \\
Random ($n=5{,}648$) & $86.7$ & $46.5$ & $49.0$ & $52.3$ & $52.3$ & $51.3$ & $43.8$ & $25.1$ \\
Cleaned ($n=5{,}648$) & $90.7$ & $46.5$ & $\mathbf{52.9}$ & $\mathbf{53.9}$ & $\mathbf{54.3}$ & $\mathbf{54.0}$ & $\mathbf{45.6}$ & $\mathbf{28.1}$ \\
\bottomrule
\end{tabular}%
}
\end{table}

\paragraph{Results.}
The cleaned set beats the random control at every epoch (Table~\ref{tab:sft_wash_ablation}), by $2.6$ points on average ($+2.0$ at Epoch 3). The four epochs come from one training run per set, but each gap is $1.2$--$3\times$ the measured run-to-run noise of about $1.3$ points, and all four point the same way. The random set sits between the cleaned and bad-kept sets at Epochs 2--4, a dose response that matches their BHI order. The gaps grow with difficulty. Most removals are per-entry bugs, so this experiment does not separate consistency from correctness; the next one does.

\subsection{In-Context Examples That Agree With Each Other}
\label{sec:p3_experiment}
Some entries are not wrong but follow a different convention from their neighbors, as when \texttt{COUNT(id)} and \texttt{COUNT(DISTINCT id)} both answer ``how many accounts.'' Does it matter whether the examples in the prompt agree with each other?

\paragraph{Test set.}
From an earlier run of Track~B (§\ref{sec:two_track}) we take $84$ pairs of real BIRD questions that the adjudicator settled by a rule, in which both entries are judged defensible, the two SQL return different results, and neither question settles the convention in words (Appendix~\ref{app:p3_details}): $53$ settled by R1 (\texttt{COUNT(DISTINCT)} vs.\ \texttt{COUNT}), $16$ by R6 (\texttt{LIMIT 1} vs.\ \texttt{MAX}), and $15$ by other rules. The realization the rule prefers is convention~A and the other is convention~B. Gemini-3.8-Flash answers $120$ exam questions from these pairs at temperature~0 under four conditions: \textbf{no context}; \textbf{matched}, examples that follow the same convention as the question's gold SQL; \textbf{opposite}, examples that follow the other convention; and \textbf{mixed}, a roughly equal number of each, which is what an uncleaned bank supplies. Each prompt holds about $17$ examples from other pairs, one per database, never the question's own pair.

\paragraph{Outside questions.}
Exam questions are twins of the examples, so we also test $151$ BIRD counting questions that belong to no pair and whose wording leaves the counting grain open, answered with the $14$ R1 examples (Appendix~\ref{app:p3_details}).

\begin{table}[t]
\centering
\caption{Execution accuracy (\%) of Gemini-3.8-Flash on convention-open BIRD questions under four context conditions. \emph{Outside questions} belong to no pair and use the R1 examples. The bottom row is matched minus opposite (exact McNemar test); Table~\ref{tab:p3_mirror} splits it by convention.}
\label{tab:p3_results}
\footnotesize
\begin{tabular}{lcc}
\toprule
\textbf{Context in prompt} & \textbf{Paired questions} & \textbf{Outside questions} \\
 & ($n = 120$) & ($n = 151$) \\
\midrule
Matched convention & $\mathbf{66.7}$ & $\mathbf{70.2}$ \\
Mixed (uncleaned bank) & $60.8$ & $62.9$ \\
None (zero-shot) & $58.3$ & $66.2$ \\
Opposite convention & $\mathbf{53.3}$ & $\mathbf{40.4}$ \\
\midrule
Matched $-$ opposite & $+13.3$ ($p = 0.0004$) & $+29.8$ ($p < 10^{-11}$) \\
\bottomrule
\end{tabular}
\end{table}

\paragraph{Results.}
Consistent examples help and conflicting ones hurt (Table~\ref{tab:p3_results}). On the paired questions, matched examples reach $66.7\%$ and opposite ones $53.3\%$, a gap of $13.3$ points; of the $20$ questions whose answer changed between the two conditions, $18$ moved in the predicted direction (exact McNemar test, $p = 0.0004$). The effect reaches beyond the bank's own questions: on the $151$ outside questions, matched examples beat opposite ones by $29.8$ points ($47$ vs.\ $2$ flips, $p = 4 \times 10^{-12}$) and mixed ones by $7.3$ ($p = 0.013$), so examples from other databases move the model on questions none of them resemble. Mixed examples stay within $3.5$ points of no examples, and matched beats opposite whichever side the gold follows (Appendix~\ref{app:p3_details}). BHI scores the matched and opposite sets as equally clean: which convention to agree on is the organization's choice, recorded in its rulebook. In a replication on $470$ naturally grouped BIRD questions, with no judge or scorer involved, consistent examples beat mixed ones by $12.1$ points where the wording leaves the convention open and by only $0.8$ where it fixes it (Appendix~\ref{app:p3_details}).

\paragraph{Dose response on a synthesized bank.} We also synthesize a bank in which every entry is judged defensible and we control the share that follows the opposite counting or superlative convention from the test gold (Appendix~\ref{app:badbank}). On $246$ held-out real BIRD and Spider questions, given the $8$ most similar entries, Gemini-3.8-Flash falls from $75.2\%$ with a consistent bank to $58.1\%$ when half the bank conflicts, against $57.3\%$ with no context, and Claude-Sonnet-5 from $70.7\%$ to $54.1\%$ ($49.2\%$ with no context). Accuracy drops at every step; $44$ questions go from right to wrong for each model against $2$ or $3$ the other way ($p < 10^{-9}$), and BHI falls with accuracy (Spearman $0.90$; Table~\ref{tab:badbank}).

\section{CAT: Measuring How Far a System Generalizes from Its Context Bank}
\label{sec:cat_benchmark}

Once a system has absorbed a clean bank, through retrieval or fine-tuning, how far beyond the bank's own questions does that knowledge reach? The ACN profile supplies the needed distance: $\alpha \approx 1$ is a rewording, $\alpha \approx 0.5$ shares about half the decisions, and $\alpha = 0$ is unrelated. \textbf{CAT} uses this to \emph{grow} a clean bank with new questions at chosen distances, verifying with the scorer that each addition sits where intended and conflicts with none of its nearest bank entries, and then to \emph{test} a system as questions move away from the bank. It tests \emph{recall} of rephrased bank questions ($d_0$), \emph{generalization} as a question shares less of its logic with the bank ($d_1$--$d_3$), and \emph{robustness} on databases the bank does not cover ($d_{\max}$); a system that uses its bank well should be near ceiling at $d_0$ and fall step by step toward $d_{\max}$.

\subsection{Generating Questions at a Controlled Distance}
CAT starts from a cleaned bank and held-out questions from its databases that serve only as a source of unseen logic. The CAT-BIRD bank holds $1{,}501$ entries from $14$ BIRD databases, six from BIRD-dev and eight from BIRD-train, with $559$ questions held out; the CAT-Spider bank holds $1{,}833$ entries from $42$ Spider-train databases, with $712$ held out. Each test question, or \emph{probe}, is made by having an LLM rewrite a bank question so that a target fraction of its logic is new, borrowed from a held-out question of the same database or written as a plausible database convention, in the style of the source benchmark; Gemini-3.8-Flash and Claude-Sonnet-5 share the generation. At $d_0$ the LLM only rephrases a bank question and the SQL is unchanged; the scorer must confirm that the rephrase preserves the entry ($\hat{\alpha} \ge 0.70$) and conflicts with no bank entry. At $d_1$ ($\hat{\alpha} \in [0.55, 0.85)$), $d_2$ ($[0.35, 0.55)$), and $d_3$ ($[0.15, 0.35)$), roughly a quarter, a half, and three quarters of the logic is new. A probe must pass five quality gates, including agreement of two LLMs that its gold SQL answers the question and an in-band $\hat{\alpha}$ with no detected conflict (Appendix~\ref{app:cat_build}). The $d_1$--$d_3$ probes are drawn from those that pass so that their no-context accuracy matches $d_0$, using only no-context scores, never those of a context-aware system. The $d_{\max}$ group ($\hat{\alpha} \approx 0$) holds real questions from databases the bank never covers: five BIRD databases for CAT-BIRD, matched to $d_0$ in no-context accuracy, and Spider-dev for CAT-Spider, matched in hardness levels. CAT-BIRD has $1{,}097$ questions and CAT-Spider $953$ (group sizes in Tables~\ref{tab:cat_results} and~\ref{tab:cat_spider}); CAT-Spider's $d_2$ and $d_3$ fell short of their targets and are harder without context than its $d_0$.

\subsection{Evaluating Context-Aware Systems on CAT-BIRD and CAT-Spider}
We evaluate two kinds of context-aware system. \emph{Retrieval}: Gemini-3.8-Flash and Claude-Sonnet-5 at temperature~$0$, given the schema alone or the schema plus the three bank entries with the most word overlap from the same database (for $d_{\max}$, from the whole bank). \emph{Fine-tuning}: Qwen2.5-7B-Instruct and Gemma-4-12B-it, fine-tuned with LoRA on the $1{,}501$ bank entries for $20$ epochs, until they reproduce the bank, and then prompted with the schema and question only, so that any bank knowledge must sit in the weights. For Qwen we also train a \emph{no-bank control} with the same recipe on $1{,}501$ cleaned BIRD-train entries from $14$ other databases, matched to the bank on per-database counts, prompt length, and query shape, which separates what the bank teaches from generic practice at BIRD-style SQL (Appendix~\ref{app:cat_sft}).

\begin{table}[t]
\centering
\caption{Execution accuracy (\%) on CAT-BIRD by distance between the test question and the context bank, with group sizes. Parentheses give the gain from the bank: retrieval over no context, and bank fine-tuning over the no-bank control (Qwen) or the base model (Gemma). Fine-tuned rows: epoch $20$, no retrieved context (Appendix~\ref{app:cat_sft}).}
\label{tab:cat_results}
\footnotesize
\setlength{\tabcolsep}{5pt}
\resizebox{\textwidth}{!}{%
\begin{tabular}{llccccc}
\toprule
\textbf{Model} & \textbf{System} & $d_0$ & $d_1$ & $d_2$ & $d_3$ & $d_{\max}$ \\
\midrule
\multicolumn{2}{l}{\textbf{CAT-BIRD} ($n$ per group)} & $397$ & $200$ & $200$ & $100$ & $200$ \\
\midrule
\multirow{2}{*}{Gemini-3.8-Flash} & No context & $76.8$ & $78.0$ & $76.5$ & $76.0$ & $74.0$ \\
 & Top-3 retrieved & $\mathbf{99.0}$\gain{+22.2} & $\mathbf{90.0}$\gain{+12.0} & $\mathbf{87.0}$\gain{+10.5} & $\mathbf{79.0}$\gain{+3.0} & $\mathbf{77.0}$\gain{+3.0} \\
\midrule
\multirow{2}{*}{Claude-Sonnet-5} & No context & $75.8$ & $76.5$ & $78.5$ & $78.0$ & $71.0$ \\
 & Top-3 retrieved & $96.7$\gain{+20.9} & $\mathbf{90.0}$\gain{+13.5} & $86.0$\gain{+7.5} & $77.0$\gain{-1.0} & $76.0$\gain{+5.0} \\
\midrule
\multirow{3}{*}{Qwen2.5-7B} & Base (no context) & $48.4$ & $35.5$ & $40.0$ & $37.0$ & $33.5$ \\
 & No-bank control SFT & $54.4$ & $39.5$ & $36.5$ & $40.0$ & $46.0$ \\
 & Bank SFT & $92.4$\gain{+38.0} & $55.0$\gain{+15.5} & $45.5$\gain{+9.0} & $41.0$\gain{+1.0} & $48.0$\gain{+2.0} \\
\midrule
\multirow{2}{*}{Gemma-4-12B} & Base (no context) & $71.8$ & $67.5$ & $73.0$ & $65.0$ & $69.0$ \\
 & Bank SFT & $98.0$\gain{+26.2} & $75.0$\gain{+7.5} & $58.0$\gain{-15.0} & $62.0$\gain{-3.0} & $57.0$\gain{-12.0} \\
\bottomrule
\end{tabular}%
}
\end{table}

\paragraph{The value of retrieval decays with distance.}
Retrieval is worth $22.2$ points to Gemini and $20.9$ to Claude at $d_0$, taking both near ceiling ($99.0\%$ and $96.7\%$). The gain then falls to $12.0$--$13.5$ points at $d_1$, $7.5$--$10.5$ at $d_2$, and $3$ or less at $d_3$; by $d_3$ the bank is worth no more than on $d_{\max}$, whose databases it does not cover at all. Without the bank, accuracy is flat along the ladder ($76.8 \to 76.0\%$ for Gemini, $75.8 \to 78.0\%$ for Claude), a flatness imposed by selection (Appendix~\ref{app:limitations}), so the decay reflects how well each model exploits an increasingly distant example.

\paragraph{Fine-tuning fades with distance too.}
Both fine-tuned models learn the bank ($d_0$). Against its control, Qwen gains $15.5$ points at $d_1$, $9.0$ at $d_2$, and $1.0$ at $d_3$, close to the retrieval gains. Gemma, trained at a higher learning rate, pays for its near-perfect recall: it ends $15.0$ points below its own base at $d_2$ and $12.0$ below at $d_{\max}$ (paired $95\%$ intervals exclude zero), losing $d_{\max}$ accuracy as $d_0$ climbs (Table~\ref{tab:cat_sft_epochs}).

\paragraph{Every system shows the expected shape on CAT-BIRD.} For all four systems, accuracy is highest at $d_0$ and falls as the question moves away (Gemma's $d_2$ and $d_3$ are within noise); retrieval and fine-tuning order the groups the same way, evidence that the ladder measures context generalization rather than something specific to retrieval. An irrelevant bank does not hurt retrieval ($+3.0$ and $+5.0$ on $d_{\max}$).

\paragraph{CAT-Spider.} Retrieval gains shrink from $14.3$ and $18.5$ points at $d_0$ to $2.0$ and $6.7$ at $d_2$, and neither $d_3$ interval excludes zero (Table~\ref{tab:cat_spider} in Appendix~\ref{app:cat_spider}). Against a no-bank control trained on other Spider-train databases, Qwen fine-tuned on the bank gains $22.9$, $14.0$, and $8.7$ points at $d_0$--$d_2$ but falls below the control at $d_3$ and $d_{\max}$, so on this instance fine-tuning does not follow the expected shape beyond $d_2$.

\section{Conclusion}
\label{sec:conclusion}
Context-aware Text-to-SQL raises two questions that per-entry review cannot answer: whether a bank agrees with itself, and whether a system that uses it generalizes from it. We answered both by comparing queries decision by decision. Turned inward, the comparison makes context bank rot measurable and repairable: a cleaned bank beats a same-size random subset at every epoch of fine-tuning, its removals agree with an independent expert audit, and in-context examples that follow one convention beat the opposite one by $13.3$ points even though every entry is defensible, with accuracy falling steadily as conflicting examples are added. Turned outward, it lets CAT expand a bank at a controlled distance, showing that the benefit of a bank falls step by step with distance and that fine-tuning hard enough to reproduce the bank can cost accuracy on databases it does not cover.

\paragraph{Limitations.}
Several caveats bound these claims (Appendix~\ref{app:limitations}). The scorer is distilled mostly on SQLite, and its thresholds may need recalibration per dialect. Cleaning, convention labels, and the correctness of generated questions rely on LLM referees rather than expert annotation, though our removals agree closely with an expert audit. The training experiment measures cleaning as a whole, not consistency alone. On CAT, the flat no-context baseline is imposed by selection and is not reached on CAT-Spider's farthest groups, the probes were generated by two of the evaluated models, and $d_3$ is small. Each experiment uses one seed per configuration.

\subsubsection*{Reproducibility Statement}
All components needed to reproduce our results are specified in the paper: the semantic unit schema and licensing principle (§\ref{sec:semantic_foundation}), the judge abstention policy (Eq.~\ref{eq:consensus_policy}), the scorer's training mix, input format, and hyperparameters (§\ref{sec:scorer_distillation} and Appendix~\ref{app:scorer}), the Bank Health Index (§\ref{sec:bank_health_index}), the two-track cleaning protocol and its rulebook (§\ref{sec:two_track}, Appendix~\ref{app:wash}), the construction of the size- and distribution-matched control arm (§\ref{sec:hygiene_impact}), the convention-open selection rules for the in-context experiment (§\ref{sec:p3_experiment}), and the probe generation and quality gates of CAT (§\ref{sec:cat_benchmark}). All experiments derive from the public BIRD, Spider, and LiveSQLBench corpora. Appendix~\ref{app:calibration} reports the tier-by-tier calibration audit with verbatim judge traces, Appendix~\ref{app:cat_sft} reports the fine-tuning recipes and per-epoch CAT-BIRD results, and Appendix~\ref{app:p3_details} reports the in-context protocol. We will release the CAT pipeline, CAT-BIRD and CAT-Spider with their context banks, the synthesized bank of Appendix~\ref{app:badbank}, the distilled scorer weights, and the per-entry cleaning ledgers with their keep and drop rationales.

\subsubsection*{Ethics Statement}
This work uses only public, non-personal Text-to-SQL research corpora and introduces no new human-subjects data. Its practical aim, removing contradictory and buggy supervision from curated query banks, is intended to reduce silently incorrect analytics results. We note one dual-use consideration: because our method demonstrates that a consistent context bank steers model behavior strongly, the same mechanism could be used to enforce a convention that is self-consistent yet substantively wrong. Bank conventions should therefore be treated as auditable organizational policy, which is precisely what the per-entry cleaning ledger is designed to make explicit.

\subsubsection*{AI Use Statement}
In this work, we used generative AI tools for prose editing and \LaTeX{} formatting. We have not used generative AI tools for research ideation, experimental design, data analysis, or producing any reported number; every number comes from the pipelines described in the paper. Large language models are also an object of study, and each such use is documented in the main text: they serve as the human-calibrated judge system for ACN profiles (§\ref{sec:scorer_distillation}), as auditors and adjudicator in both cleaning tracks (§\ref{sec:two_track}), as probe generators and correctness judges in CAT (§\ref{sec:cat_benchmark}), as twin writers for the synthesized bank (Appendix~\ref{app:badbank}), and as the evaluated systems (§\ref{sec:hygiene_impact}, §\ref{sec:cat_benchmark}). We have reviewed all AI-assisted work, and we take responsibility for the final content of this work, including text, claims, and artifacts produced with the aid of generative AI.

\clearpage
\bibliography{iclr2027_conference}

\begin{thebibliography}{50}
\providecommand{\natexlab}[1]{#1}
\providecommand{\url}[1]{\texttt{#1}}
\expandafter\ifx\csname urlstyle\endcsname\relax
  \providecommand{\doi}[1]{doi: #1}\else
  \providecommand{\doi}{doi: \begingroup \urlstyle{rm}\Url}\fi

\bibitem[Arslan et~al.(2024)Arslan, Ghanem, Munawar, and
  Cruz]{arslan2024survey}
Muhammad Arslan, Hussam Ghanem, Saba Munawar, and Christophe Cruz.
\newblock A survey on rag with llms.
\newblock \emph{Procedia computer science}, 246:\penalty0 3781--3790, 2024.

\bibitem[Bhaskar et~al.(2023)Bhaskar, Tomar, Sathe, and
  Sarawagi]{bhaskar2023ambiqt}
Adithya Bhaskar, Tushar Tomar, Ashutosh Sathe, and Sunita Sarawagi.
\newblock Benchmarking and improving text-to-{SQL} generation under ambiguity.
\newblock In \emph{Proceedings of the 2023 Conference on Empirical Methods in
  Natural Language Processing}, pp.\  7053--7074, 2023.

\bibitem[Brown et~al.(2020)Brown, Mann, Ryder, Subbiah, Kaplan, Dhariwal,
  Neelakantan, Shyam, Sastry, Askell, et~al.]{brown2020language}
Tom Brown, Benjamin Mann, Nick Ryder, Melanie Subbiah, Jared~D Kaplan, Prafulla
  Dhariwal, Arvind Neelakantan, Pranav Shyam, Girish Sastry, Amanda Askell,
  et~al.
\newblock Language models are few-shot learners.
\newblock \emph{Advances in neural information processing systems},
  33:\penalty0 1877--1901, 2020.

\bibitem[Chang et~al.(2023)Chang, Wang, Dong, Pan, Zhu, Li, Lan, Zhang, Jiang,
  Lilien, et~al.]{chang2023dr}
Shuaichen Chang, Jun Wang, Mingwen Dong, Lin Pan, Henghui Zhu, Alexander~Hanbo
  Li, Wuwei Lan, Sheng Zhang, Jiarong Jiang, Joseph Lilien, et~al.
\newblock Dr. spider: A diagnostic evaluation benchmark towards text-to-sql
  robustness.
\newblock \emph{arXiv preprint arXiv:2301.08881}, 2023.

\bibitem[Chen et~al.(2023)Chen, Lin, Sch{\"a}rli, and Zhou]{chen2023teaching}
Xinyun Chen, Maxwell Lin, Nathanael Sch{\"a}rli, and Denny Zhou.
\newblock Teaching large language models to self-debug.
\newblock \emph{arXiv preprint arXiv:2304.05128}, 2023.

\bibitem[Deng et~al.(2021)Deng, Awadallah, Meek, Polozov, Sun, and
  Richardson]{deng2021structure}
Xiang Deng, Ahmed~Hassan Awadallah, Christopher Meek, Oleksandr Polozov, Huan
  Sun, and Matthew Richardson.
\newblock Structure-grounded pretraining for text-to-{SQL}.
\newblock In \emph{Proceedings of the 2021 Conference of the North American
  Chapter of the Association for Computational Linguistics: Human Language
  Technologies}, pp.\  1337--1350, 2021.

\bibitem[Dong et~al.(2025)Dong, Kumar, Hu, Chauhan, Hang, Chang, Pan, Lan, Zhu,
  Jiang, Ng, and Wang]{dong2025practiq}
Mingwen Dong, Nischal~Ashok Kumar, Yiqun Hu, Anuj Chauhan, Chung-Wei Hang,
  Shuaichen Chang, Lin Pan, Wuwei Lan, Henghui Zhu, Jiarong Jiang, Patrick Ng,
  and Zhiguo Wang.
\newblock {PRACTIQ}: A practical conversational text-to-{SQL} dataset with
  ambiguous and unanswerable queries.
\newblock In \emph{Proceedings of the 2025 Conference of the North American
  Chapter of the Association for Computational Linguistics}, pp.\  255--273,
  2025.

\bibitem[Entezari~Maleki et~al.(2025)Entezari~Maleki, Pourreza, and
  Rafiei]{entezari2025confidence}
Sepideh Entezari~Maleki, Mohammadreza Pourreza, and Davood Rafiei.
\newblock Confidence estimation for text-to-{SQL} in large language models.
\newblock \emph{arXiv preprint arXiv:2508.14056}, 2025.

\bibitem[Gan et~al.(2021{\natexlab{a}})Gan, Chen, Huang, Purver, Woodward, Xie,
  and Huang]{gan2021towards}
Yujian Gan, Xinyun Chen, Qiuping Huang, Matthew Purver, John~R. Woodward,
  Jinxia Xie, and Pengsheng Huang.
\newblock Towards robustness of text-to-{SQL} models against synonym
  substitution.
\newblock In \emph{Proceedings of the 59th Annual Meeting of the Association
  for Computational Linguistics and the 11th International Joint Conference on
  Natural Language Processing (Volume 1: Long Papers)}, pp.\  2505--2515,
  2021{\natexlab{a}}.

\bibitem[Gan et~al.(2021{\natexlab{b}})Gan, Chen, and Purver]{gan2021exploring}
Yujian Gan, Xinyun Chen, and Matthew Purver.
\newblock Exploring underexplored limitations of cross-domain text-to-sql
  generalization.
\newblock \emph{arXiv preprint arXiv:2109.05157}, 2021{\natexlab{b}}.

\bibitem[Gan et~al.(2022)Gan, Chen, Huang, and Purver]{gan2022measuring}
Yujian Gan, Xinyun Chen, Qiuping Huang, and Matthew Purver.
\newblock Measuring and improving compositional generalization in text-to-{SQL}
  via component alignment.
\newblock In \emph{Findings of the Association for Computational Linguistics:
  NAACL 2022}, pp.\  831--843, 2022.

\bibitem[Gao et~al.(2023)Gao, Wang, Li, Sun, Qian, Ding, and Zhou]{gao2023text}
Dawei Gao, Haibin Wang, Yaliang Li, Xiuyu Sun, Yichen Qian, Bolin Ding, and
  Jingren Zhou.
\newblock Text-to-sql empowered by large language models: A benchmark
  evaluation.
\newblock \emph{arXiv preprint arXiv:2308.15363}, 2023.

\bibitem[Gao et~al.(2024)Gao, Wang, Li, Sun, Qian, Ding, and
  Zhou]{gao2024dailsql}
Dawei Gao, Haibin Wang, Yongbin Li, Xiangnan Sun, Yong Qian, Bolin Ding, and
  Jingren Zhou.
\newblock {DAIL-SQL}: Optimizing text-to-{SQL} via diverse examples, 2024.

\bibitem[Hu et~al.(2021)Hu, Shen, Wallis, Allen-Zhu, Li, Wang, Wang, and
  Chen]{hu2021lora}
Edward~J Hu, Yelong Shen, Phillip Wallis, Zeyuan Allen-Zhu, Yuanzhi Li, Shean
  Wang, Lu~Wang, and Weizhu Chen.
\newblock Lora: Low-rank adaptation of large language models.
\newblock \emph{arXiv preprint arXiv:2106.09685}, 2021.

\bibitem[Iyer et~al.(2017)Iyer, Konstas, Cheung, Krishnamurthy, and
  Zettlemoyer]{iyer2017learning}
Srinivasan Iyer, Ioannis Konstas, Alvin Cheung, Jayant Krishnamurthy, and Luke
  Zettlemoyer.
\newblock Learning a neural semantic parser from user feedback.
\newblock \emph{arXiv preprint arXiv:1704.08760}, 2017.

\bibitem[Jin et~al.(2026)Jin, Choi, Zhu, and Kang]{jin2026annotation}
Tianyi Jin, Yejin Choi, Yu~Zhu, and Daniel Kang.
\newblock Text-to-{SQL} benchmarks are broken: An in-depth analysis of
  annotation errors.
\newblock In \emph{Conference on Innovative Data Systems Research (CIDR)},
  2026.

\bibitem[Keysers et~al.(2020)Keysers, Sch{\"a}rli, Scales, Buisman, Furrer,
  Kashubin, Momchev, Sinopalnikov, Stafiniak, Tihon, Tsarkov, Wang, van Zee,
  and Bousquet]{keysers2020measuring}
Daniel Keysers, Nathanael Sch{\"a}rli, Nathan Scales, Hylke Buisman, Daniel
  Furrer, Sergii Kashubin, Nikola Momchev, Danila Sinopalnikov, Lukasz
  Stafiniak, Tibor Tihon, Dmitry Tsarkov, Xiao Wang, Marc van Zee, and Olivier
  Bousquet.
\newblock Measuring compositional generalization: A comprehensive method on
  realistic data.
\newblock In \emph{International Conference on Learning Representations}, 2020.

\bibitem[Khot et~al.(2022)Khot, Trivedi, Finlayson, Fu, Richardson, Clark, and
  Sabharwal]{khot2022decomposed}
Tushar Khot, Harsh Trivedi, Matthew Finlayson, Yao Fu, Kyle Richardson, Peter
  Clark, and Ashish Sabharwal.
\newblock Decomposed prompting: A modular approach for solving complex tasks.
\newblock \emph{arXiv preprint arXiv:2210.02406}, 2022.

\bibitem[Kuhn et~al.(2023)Kuhn, Gal, and Farquhar]{kuhn2023semantic}
Lorenz Kuhn, Yarin Gal, and Sebastian Farquhar.
\newblock Semantic uncertainty: Linguistic invariances for uncertainty
  estimation in natural language generation.
\newblock In \emph{International Conference on Learning Representations}, 2023.

\bibitem[Lee et~al.(2025)Lee, Yim, Shin, Song, Kim, and Byun]{lee2025mcssql}
Dongjun Lee, Choongwon Yim, Jiho Shin, Hyeongrae Song, Sanghoon Kim, and
  Jaehyuk Byun.
\newblock {MCS-SQL}: Leveraging multiple prompts and multiple-choice selection
  for text-to-{SQL} generation, 2025.

\bibitem[Lei et~al.(2020)Lei, Wang, Ma, Gan, Lu, Kan, and Chua]{lei2020re}
Wenqiang Lei, Weixin Wang, Zhixin Ma, Tian Gan, Wei Lu, Min-Yen Kan, and
  Tat-Seng Chua.
\newblock Re-examining the role of schema linking in text-to-sql.
\newblock In \emph{Proceedings of the 2020 Conference on Empirical Methods in
  Natural Language Processing (EMNLP)}, pp.\  6943--6954, 2020.

\bibitem[Li et~al.(2023{\natexlab{a}})Li, Zhang, Li, and
  Chen]{li2023decoupling}
Haoyang Li, Jing Zhang, Cuiping Li, and Hong Chen.
\newblock Decoupling the skeleton parsing and schema linking for text-to-sql.
\newblock \emph{arXiv preprint arXiv:2302.05965}, 2023{\natexlab{a}}.

\bibitem[Li et~al.(2023{\natexlab{b}})Li, Zhang, Li, and Chen]{li2023resdsql}
Haoyang Li, Jing Zhang, Cuiping Li, and Hong Chen.
\newblock Resdsql: Decoupling schema linking and skeleton parsing for
  text-to-sql.
\newblock In \emph{Proceedings of the AAAI Conference on Artificial
  Intelligence}, volume~37, pp.\  13067--13075, 2023{\natexlab{b}}.

\bibitem[Li et~al.(2024{\natexlab{a}})Li, Zhang, Liu, Fan, Zhang, Zhu, Wei,
  Pan, Li, and Chen]{li2024codes}
Haoyang Li, Jing Zhang, Hanbing Liu, Ju~Fan, Xiaokang Zhang, Jun Zhu, Renjie
  Wei, Hongyan Pan, Cuiping Li, and Hong Chen.
\newblock Codes: Towards building open-source language models for text-to-sql.
\newblock \emph{Proceedings of the ACM on Management of Data}, 2\penalty0
  (3):\penalty0 1--28, 2024{\natexlab{a}}.

\bibitem[Li et~al.(2025)Li, Wu, Zhang, Huang, Zhang, Jiang, Wang, Zhang, Chen,
  Shi, Chen, and Li]{li2025omnisql}
Haoyang Li, Shang Wu, Xiaokang Zhang, Xinmei Huang, Jing Zhang, Fuxin Jiang,
  Shuai Wang, Tieying Zhang, Jianjun Chen, Rui Shi, Hong Chen, and Cuiping Li.
\newblock {OmniSQL}: Synthesizing high-quality text-to-{SQL} data at scale.
\newblock \emph{Proceedings of the VLDB Endowment}, 2025.

\bibitem[Li et~al.(2024{\natexlab{b}})Li, Hui, Qu, Yang, Li, Li, Wang, Qin,
  Geng, Huo, et~al.]{li2024can}
Jinyang Li, Binyuan Hui, Ge~Qu, Jiaxi Yang, Binhua Li, Bowen Li, Bailin Wang,
  Bowen Qin, Ruiying Geng, Nan Huo, et~al.
\newblock Can llm already serve as a database interface? a big bench for
  large-scale database grounded text-to-sqls.
\newblock \emph{Advances in Neural Information Processing Systems}, 36,
  2024{\natexlab{b}}.

\bibitem[Liu et~al.(2026)Liu, Zhu, Gao, Luo, Li, Shi, Hong, Gao, Li, Ding,
  et~al.]{liu2026xiyan}
Yifu Liu, Yin Zhu, Yingqi Gao, Zhiling Luo, Xiaoxia Li, Xiaorong Shi, Yuntao
  Hong, Jinyang Gao, Yu~Li, Bolin Ding, et~al.
\newblock Xiyan-sql: A novel multi-generator framework for text-to-sql.
\newblock \emph{IEEE Transactions on Knowledge and Data Engineering}, 2026.

\bibitem[Luoma \& Kumar(2025)Luoma and Kumar]{luoma2025snails}
Kyle Luoma and Arun Kumar.
\newblock {SNAILS}: Schema naming assessments for improved {LLM}-based {SQL}
  inference.
\newblock \emph{Proceedings of the ACM on Management of Data (SIGMOD)},
  3\penalty0 (1), 2025.

\bibitem[Ma et~al.(2026)Ma, Zhuang, Xu, Jiang, Chen, and Guo]{ma2026sql}
Peixian Ma, Xialie Zhuang, Chengjin Xu, Xuhui Jiang, Ran Chen, and Jian Guo.
\newblock Sql-r1: Training natural language to sql reasoning model by
  reinforcement learning.
\newblock \emph{Advances in Neural Information Processing Systems},
  38:\penalty0 174505--174537, 2026.

\bibitem[Ni et~al.(2023)Ni, Iyer, Radev, Stoyanov, Yih, Wang, and
  Lin]{ni2023lever}
Ansong Ni, Srini Iyer, Dragomir Radev, Veselin Stoyanov, Wen-tau Yih, Sida
  Wang, and Xi~Victoria Lin.
\newblock Lever: Learning to verify language-to-code generation with execution.
\newblock In \emph{International Conference on Machine Learning}, pp.\
  26106--26128. PMLR, 2023.

\bibitem[Pourreza \& Rafiei(2023)Pourreza and Rafiei]{pourreza2023evaluating}
Mohammadreza Pourreza and Davood Rafiei.
\newblock Evaluating cross-domain text-to-sql models and benchmarks.
\newblock \emph{arXiv preprint arXiv:2310.18538}, 2023.

\bibitem[Pourreza \& Rafiei(2024{\natexlab{a}})Pourreza and
  Rafiei]{pourreza2024din}
Mohammadreza Pourreza and Davood Rafiei.
\newblock Din-sql: Decomposed in-context learning of text-to-sql with
  self-correction.
\newblock \emph{Advances in Neural Information Processing Systems}, 36,
  2024{\natexlab{a}}.

\bibitem[Pourreza \& Rafiei(2024{\natexlab{b}})Pourreza and
  Rafiei]{pourreza2024dts}
Mohammadreza Pourreza and Davood Rafiei.
\newblock Dts-sql: Decomposed text-to-sql with small large language models.
\newblock \emph{arXiv preprint arXiv:2402.01117}, 2024{\natexlab{b}}.

\bibitem[Pourreza et~al.(2025{\natexlab{a}})Pourreza, Li, Sun, Chung, Talaei,
  Kakkar, Gan, Saberi, Ozcan, and Arik]{pourreza2025chase}
Mohammadreza Pourreza, Hailong Li, Ruoxi Sun, Yeounoh Chung, Shayan Talaei,
  Gaurav~Tarlok Kakkar, Yu~Gan, Amin Saberi, Fatma Ozcan, and Sercan Arik.
\newblock Chase-sql: Multi-path reasoning and preference optimized candidate
  selection in text-to-sql.
\newblock In \emph{International Conference on Learning Representations},
  volume 2025, pp.\  60385--60415, 2025{\natexlab{a}}.

\bibitem[Pourreza et~al.(2025{\natexlab{b}})Pourreza, Talaei, Sun, Wan, Li,
  Mirhoseini, Saberi, Arik, et~al.]{pourreza2025reasoning}
Mohammadreza Pourreza, Shayan Talaei, Ruoxi Sun, Xingchen Wan, Hailong Li,
  Azalia Mirhoseini, Amin Saberi, Sercan Arik, et~al.
\newblock Reasoning-sql: Reinforcement learning with sql tailored partial
  rewards for reasoning-enhanced text-to-sql.
\newblock \emph{arXiv preprint arXiv:2503.23157}, 2025{\natexlab{b}}.

\bibitem[Shaw et~al.(2021)Shaw, Chang, Pasupat, and
  Toutanova]{shaw2021compositional}
Peter Shaw, Ming-Wei Chang, Panupong Pasupat, and Kristina Toutanova.
\newblock Compositional generalization and natural language variation: Can a
  semantic parsing approach handle both?
\newblock In \emph{Proceedings of the 59th Annual Meeting of the Association
  for Computational Linguistics and the 11th International Joint Conference on
  Natural Language Processing (Volume 1: Long Papers)}, pp.\  922--938, 2021.

\bibitem[Shinn et~al.(2024)Shinn, Cassano, Gopinath, Narasimhan, and
  Yao]{shinn2024reflexion}
Noah Shinn, Federico Cassano, Ashwin Gopinath, Karthik Narasimhan, and Shunyu
  Yao.
\newblock Reflexion: Language agents with verbal reinforcement learning.
\newblock \emph{Advances in Neural Information Processing Systems}, 36, 2024.

\bibitem[Shkapenyuk et~al.(2025)Shkapenyuk, Srivastava, Johnson, and
  Ghane]{shkapenyuk2025automatic}
Vladislav Shkapenyuk, Divesh Srivastava, Theodore Johnson, and Parisa Ghane.
\newblock Automatic metadata extraction for text-to-sql.
\newblock \emph{arXiv preprint arXiv:2505.19988}, 2025.

\bibitem[Sun et~al.(2023)Sun, Arik, Sinha, Nakhost, Dai, Yin, and
  Pfister]{sun2023sqlprompt}
Ruoxi Sun, Sercan~{\"O} Arik, Rajarishi Sinha, Hootan Nakhost, Hanjun Dai,
  Pengcheng Yin, and Tomas Pfister.
\newblock Sqlprompt: In-context text-to-sql with minimal labeled data.
\newblock \emph{arXiv preprint arXiv:2311.02883}, 2023.

\bibitem[Talaei et~al.(2024)Talaei, Pourreza, Chang, Mirhoseini, and
  Saberi]{talaei2024chess}
Shayan Talaei, Mohammadreza Pourreza, Yu-Chen Chang, Azalia Mirhoseini, and
  Amin Saberi.
\newblock Chess: Contextual harnessing for efficient sql synthesis.
\newblock \emph{arXiv preprint arXiv:2405.16755}, 2024.

\bibitem[Vaidya et~al.(2025)Vaidya, Sankararaman, Ding, Lei, Qin,
  Narayanaswamy, and Kraska]{vaidya2025odin}
Karan Vaidya, Arun Sankararaman, Jian Ding, Chuan Lei, Xiao Qin, Balakrishnan
  Narayanaswamy, and Tim Kraska.
\newblock {ODIN}: A {NL2SQL} recommender to handle schema ambiguity, 2025.

\bibitem[Veltri et~al.(2023)Veltri, Badaro, Saeed, and
  Papotti]{veltri2023pythia}
Enzo Veltri, Gilbert Badaro, Mohammed Saeed, and Paolo Papotti.
\newblock Data ambiguity profiling for the generation of training examples.
\newblock In \emph{IEEE International Conference on Data Engineering (ICDE)},
  pp.\  450--463, 2023.

\bibitem[Wang et~al.(2024)Wang, Ren, Yang, Liang, Bai, Chai, Yan, Zhang, Yin,
  Sun, et~al.]{wang2024mac}
Bing Wang, Changyu Ren, Jian Yang, Xinnian Liang, Jiaqi Bai, Linzheng Chai,
  Zhao Yan, Qian-Wen Zhang, Di~Yin, Xing Sun, et~al.
\newblock Mac-sql: A multi-agent collaborative framework for text-to-sql.
\newblock \emph{arXiv preprint arXiv:2312.11242}, 2024.

\bibitem[Wei et~al.(2021)Wei, Bosma, Zhao, Guu, Yu, Lester, Du, Dai, and
  Le]{wei2021finetuned}
Jason Wei, Maarten Bosma, Vincent~Y Zhao, Kelvin Guu, Adams~Wei Yu, Brian
  Lester, Nan Du, Andrew~M Dai, and Quoc~V Le.
\newblock Finetuned language models are zero-shot learners.
\newblock \emph{arXiv preprint arXiv:2109.01652}, 2021.

\bibitem[Wretblad et~al.(2024)Wretblad, Riseby, Biswas, Ahmadi, and
  Holmstr{\"o}m]{wretblad2024bird}
Niklas Wretblad, Fredrik Riseby, Rahul Biswas, Amin Ahmadi, and Oskar
  Holmstr{\"o}m.
\newblock Understanding the effects of noise in text-to-{SQL}: An examination
  of the {BIRD-Bench} benchmark.
\newblock In \emph{Proceedings of the 62nd Annual Meeting of the Association
  for Computational Linguistics (Short Papers)}, 2024.

\bibitem[Yao et~al.(2026)Yao, Sun, Borchmann, Shen, Deng, Zhai, Zhang, Li, and
  He]{yao2026arctic}
Zhewei Yao, Guoheng Sun, {\L}ukasz Borchmann, Zheyu Shen, Minghang Deng, Bohan
  Zhai, Hao Zhang, Ang Li, and Yuxiong He.
\newblock Arctic-text2sql-r1: Simple rewards, strong reasoning in text-to-sql.
\newblock In \emph{Findings of the Association for Computational Linguistics:
  ACL 2026}, pp.\  26966--26995, 2026.

\bibitem[Yu et~al.(2018)Yu, Zhang, Yang, Yasunaga, Wang, Li, Ma, Li, Yao,
  Roman, et~al.]{yu2018spider}
Tao Yu, Rui Zhang, Kai Yang, Michihiro Yasunaga, Dongxu Wang, Zifan Li, James
  Ma, Irene Li, Qingning Yao, Shanelle Roman, et~al.
\newblock Spider: A large-scale human-labeled dataset for complex and
  cross-domain semantic parsing and text-to-sql task.
\newblock \emph{arXiv preprint arXiv:1809.08887}, 2018.

\bibitem[Yu et~al.(2019)Yu, Zhang, Er, Li, Xue, Pang, Lin, Tan, Shi, Li,
  et~al.]{yu2019cosql}
Tao Yu, Rui Zhang, He~Yang Er, Suyi Li, Eric Xue, Bo~Pang, Xi~Victoria Lin,
  Yi~Chern Tan, Tianze Shi, Zihan Li, et~al.
\newblock Cosql: A conversational text-to-sql challenge towards cross-domain
  natural language interfaces to databases.
\newblock \emph{arXiv preprint arXiv:1909.05378}, 2019.

\bibitem[Zhou et~al.(2022)Zhou, Sch{\"a}rli, Hou, Wei, Scales, Wang,
  Schuurmans, Bousquet, Le, and Chi]{zhou2022least}
Denny Zhou, Nathanael Sch{\"a}rli, Le~Hou, Jason Wei, Nathan Scales, Xuezhi
  Wang, Dale Schuurmans, Olivier Bousquet, Quoc Le, and Ed~Chi.
\newblock Least-to-most prompting enables complex reasoning in large language
  models.
\newblock \emph{arXiv preprint arXiv:2205.10625}, 2022.

\bibitem[Zhu et~al.(2026)Zhu, Jin, Choi, and Kang]{zhu2026revisql}
Yuxuan Zhu, Tengjun Jin, Yoojin Choi, and Daniel Kang.
\newblock Human-level text-to-sql via reinforcement learning on verified data,
  without pipeline engineering.
\newblock \emph{arXiv preprint arXiv:2603.20004}, 2026.

\end{thebibliography}
\bibliographystyle{iclr2027_conference}

\clearpage
\appendix

\section{Appendix: Related Work}
\label{sec:related_work}

\paragraph{How context banks are used.}
Context entries enter Text-to-SQL systems in two ways. At inference time they are retrieved into the prompt \citep{brown2020language, sun2023sqlprompt, gao2024dailsql, talaei2024chess, shkapenyuk2025automatic}, where they also feed decomposed reasoning \citep{zhou2022least, khot2022decomposed, pourreza2024din}, multi-agent \citep{wang2024mac}, and multi-candidate \citep{lee2025mcssql} pipelines. At training time they serve as supervision for fine-tuning \citep{wei2021finetuned, hu2021lora, pourreza2024dts, li2024codes, liu2026xiyan}, execution-guided self-correction \citep{chen2023teaching, ni2023lever, shinn2024reflexion}, preference optimization \citep{pourreza2025chase}, and reinforcement learning \citep{pourreza2025reasoning, ma2026sql, yao2026arctic}. Work on improving these pipelines has focused on selecting better examples, through retrieval similarity \citep{gao2024dailsql} or schema linking \citep{lei2020re, li2023decoupling, li2023resdsql}. We ask a prior question: whether the examples being selected from agree with one another.

\paragraph{Data quality is checked one entry at a time.}
Several audits have found annotation bugs in Text-to-SQL benchmarks \citep{wretblad2024bird, luoma2025snails, jin2026annotation}, and \citet{zhu2026revisql} showed that training on an expert-cleaned corpus (BIRD-Platinum) improves downstream accuracy. In every case the unit of inspection is a single input--SQL pair: is this SQL right for this input? A separate line of work studies natural language ambiguity in relational queries \citep{bhaskar2023ambiqt, veltri2023pythia, dong2025practiq, vaidya2025odin} and resolves it at inference time through clarification or confidence estimation \citep{iyer2017learning, kuhn2023semantic, entezari2025confidence}. We take the complementary view: the same under-specification, resolved differently by different annotators, leaves contradictions inside the bank, and those contradictions are invisible to per-entry review. Our contribution is a way to check consistency across entries, with evidence that consistency matters at inference time even when no entry is individually wrong.

\paragraph{Benchmarks do not control distance to the context.}
Standard benchmarks \citep{yu2018spider, yu2019cosql, li2024can} and cross-domain stress tests \citep{pourreza2023evaluating} evaluate a model on a fixed test set without a bank. Perturbation suites \citep{gan2021towards, gan2021exploring, deng2021structure, chang2023dr} change a question's wording or add domain knowledge, but do not grade how much new logic a question contains, so they cannot tell whether a model transfers a bank's conventions or only reproduces its examples. Compositional-generalization splits \citep{keysers2020measuring, shaw2021compositional, gan2022measuring} do quantify the distance between training and test data, but they split existing data by syntactic compound divergence rather than relating each test question to a specific bank, and they do not check the test set for conflicts with it. Large-scale synthesis \citep{li2024codes, li2025omnisql} generates new questions, but as training data and without regard to consistency with an existing bank. CAT generates new questions at a controlled semantic distance from a given bank, verifies that distance with a trained scorer, and filters out questions that conflict with the bank.

\section{Appendix: Scorer Implementation and Validation Details}
\label{app:scorer}

\paragraph{Backbone and optimization.}
The scorer is Qwen3.6-27B with a two-output sequence-classification head. We chose Qwen because, when the scorer was built, the Gemma~4 architecture had no sequence-classification head in the \texttt{transformers} library. LoRA (rank $16$, scaling $32$, dropout $0.05$) is applied to all attention and MLP projections, and the head is trained in full. We train for $2$ epochs with learning rate $2\times 10^{-5}$, cosine decay with $3\%$ warmup, effective batch size $16$, and a $3072$-token limit, which takes $30.8$ hours on one 80GB GPU. We use the final checkpoint. Scoring runs at about $5$--$9$ pairs per second on one GPU (A100 or H100, batch size $8$).

\paragraph{Frozen serialization contract.}
Inputs follow a fixed render contract
\begin{equation*}
\mathbf{x} = [\texttt{[SCHEMA]} \parallel \mathcal{D}_{\text{sub}} \parallel \texttt{[ENTRY\_A]} \parallel (Q_A, S_A) \parallel \texttt{[ENTRY\_B]} \parallel (Q_B, S_B)],
\end{equation*}
where SQLs are canonicalized via \texttt{sqlglot}, the schema block is tagged with its SQL dialect, and $\mathcal{D}_{\text{sub}}$ carries enriched value statistics for the columns touched by either query. Under the $3072$-token limit, any overflow is absorbed by the schema description only, never by the questions or the SQL, so a clause dropped by truncation can never be scored as a phantom conflict. At inference we average the two orders, $f_\theta(e_A, e_B) = \tfrac{1}{2}(f_\theta(e_A, e_B) + f_\theta(e_B, e_A))$. The two orders rarely disagree much: $\hat{\alpha}$ differs between them by $0.011$ on average ($95$th percentile $0.031$) on the $288$ calibration pairs below, and by $0.027$ ($0.074$) on $112$ real BIRD-dev pairs.

\paragraph{Accuracy.}
Table~\ref{tab:scorer_accuracy} reports the scorer on two evaluation sets. The held-out test set contains $2{,}879$ pairs from the $11$ BIRD-dev databases, which never appear in training, built by the same generators and labeled by the same rules as the training pairs. The calibration set contains the $288$ pairs from the human calibration rounds of Appendix~\ref{app:judge_calibration}: $148$ whose labels were checked or adjudicated by the human, and $140$ from the final round, of which $115$ are constructed pairs whose labels are fixed by construction and $25$ are real pairs labeled by the final judge system. These pairs carry no database reference, so they are scored without a schema fragment, a harder condition than in use. Throughout, conflict AUC treats a pair as conflicting when its label has $\beta > 0$.

\begin{table}[h]
\centering
\caption{Accuracy of the scorer. MAE and Spearman are for $\hat{\alpha}$; AUC is for $\hat{\beta}$.}
\label{tab:scorer_accuracy}
\footnotesize
\begin{tabular}{lrccc}
\toprule
\textbf{Evaluation set} & \textbf{Pairs} & \textbf{Align MAE} & \textbf{Align Spearman} & \textbf{Conflict AUC} \\
\midrule
Held-out BIRD-dev, all & $2{,}879$ & $0.020$ & $0.970$ & $0.982$ \\
\quad real pairs & $1{,}690$ & $0.020$ & $0.855$ & $0.872$ \\
\quad constructed pairs & $1{,}189$ & $0.021$ & $0.966$ & $1.000$ \\
Human calibration set & $288$ & $0.057$ & $0.934$ & $0.880$ \\
\bottomrule
\end{tabular}
\end{table}

\paragraph{Agreement with the judges.}
We compare the scorer with both judges on $112$ real BIRD-dev pairs, $71$ of which receive a consensus label ($41$ abstentions, where the judges diverge). On these $71$, the scorer's $\hat{\alpha}$ is closer to the judges' consensus (mean absolute error $0.047$, no bias, Pearson $0.84$) than the two judges are to each other ($0.053$), so it did not simply learn one judge's habits. Its $\hat{\beta}$ ranks the pairs well (AUC $0.915$ for a consensus conflict above zero, $27$ positives) but is compressed relative to the judges' (standard deviation $0.040$ vs.\ $0.084$), so a value of $0.10$ does not mean a $10\%$ chance of conflict; this is why we use $\hat{\beta}$ only to rank pairs for review.

\paragraph{Training progress.}
The calibration set was evaluated every $200$ steps during training (Table~\ref{tab:scorer_curve}). Alignment error falls steadily through the first epoch and levels off in the second; conflict AUC levels off after about $2{,}000$ steps.

\begin{table}[h]
\centering
\caption{Scorer accuracy on the human calibration set during training ($5{,}142$ steps, $2$ epochs).}
\label{tab:scorer_curve}
\footnotesize
\begin{tabular}{lcccccc}
\toprule
\textbf{Step} & $200$ & $1{,}000$ & $2{,}000$ & $3{,}000$ & $4{,}000$ & $5{,}000$ \\
\midrule
Align MAE & $0.154$ & $0.114$ & $0.067$ & $0.063$ & $0.059$ & $0.057$ \\
Align Spearman & $0.700$ & $0.878$ & $0.915$ & $0.932$ & $0.932$ & $0.934$ \\
Conflict AUC & $0.792$ & $0.848$ & $0.894$ & $0.892$ & $0.883$ & $0.880$ \\
\bottomrule
\end{tabular}
\end{table}

\paragraph{Capacity ablation.}
Training the same data with LoRA rank $256$ ($16\times$ the trainable parameters) changes held-out accuracy by at most $0.005$ on any metric (Table~\ref{tab:scorer_capacity}), so the scorer is limited by its labels rather than its capacity. An earlier version trained on the SQLite pairs only is equally accurate on this SQLite test set.

\begin{table}[h]
\centering
\caption{Held-out BIRD-dev accuracy ($2{,}879$ pairs) for scorer variants.}
\label{tab:scorer_capacity}
\footnotesize
\begin{tabular}{lccc}
\toprule
\textbf{Variant} & \textbf{Align MAE} & \textbf{Align Spearman} & \textbf{Conflict AUC} \\
\midrule
Final scorer (rank $16$, SQLite + PostgreSQL) & $0.020$ & $0.970$ & $0.982$ \\
Rank $256$, same data & $0.016$ & $0.973$ & $0.980$ \\
Earlier version (rank $128$, SQLite only) & $0.016$ & $0.972$ & $0.980$ \\
\bottomrule
\end{tabular}
\end{table}

\section{Appendix: Judge Calibration and Examples}
\label{app:calibration}

\subsection{Human Calibration of the Judge System}
\label{app:judge_calibration}

\paragraph{Protocol.}
Calibration ran as a loop. In each round we drew a batch of pairs, had both judges label them, and showed the two labels to a human expert for adjudication; the adjudications were then turned into changes to the judge instructions, to the post-processing of judge outputs, or to the pair generators, and the next round tested the changed system. The first four rounds were open: the adjudicator saw which judge produced which label, and the rounds served mainly to settle the labeling principles. The remaining rounds were blind: judge identities were hidden and randomized per pair, pairs were shuffled, and the pair type and any construction label were withheld until after adjudication. Each batch mixed synthetic variants of every type used in training, rephrases, and real pairs of entries (near-duplicates, same-question pairs, pairs that need judgment, and near-orthogonal pairs), so every part of the training distribution was reviewed. Agreement between the two judges does not guarantee that they are right, so from the fifth blind round on we also flagged, without revealing its direction, every pair where the two judges agreed with each other but not with the construction label, and the adjudicator reviewed these as well. This surfaced errors that the judges shared.

\paragraph{What the adjudications changed.}
Most changes sharpened the definitions of align, conflict, and novel. Representative rules that entered the judge instructions:
\begin{itemize}
  \item \emph{No forced alignment.} Two queries that address unrelated needs are not aligned merely because they touch the same table or share a template.
  \item \emph{Licensing decides conflict versus novel.} A decision that only one query makes has no counterpart to disagree with and is novel, following Eq.~\ref{eq:acn}. A conflict requires a role that both queries implement, realized differently in a way their questions do not explain. An extra restriction that the question states explicitly is therefore novel, not a conflict.
  \item \emph{Segmentation is not disagreement.} Two judges, or two queries, that split the same logic into units of different size do not thereby conflict.
  \item \emph{Same need, different SQL is a conflict.} Two implementations count as the same decision only if they are identical after normalization or provably interchangeable from the schema (for example \texttt{DISTINCT} over a primary key). Execution agreement on one database does not establish equivalence.
  \item \emph{Grounding.} When the two questions refer to the same stored value in forms that need knowledge to connect (an alias, an abbreviation, a unit conversion), the shared decision counts as novel rather than aligned.
\end{itemize}
Where a rule could be decided mechanically from the two queries and the schema, we moved it out of the prompt and into deterministic post-processing of the judges' unit maps, keeping the prompt for decisions that need judgment. When a judge kept failing on a rule after it was written into the prompt, we did not keep adding instructions, which would have overfit the prompt to the adjudicated batches.

\paragraph{Two judges with different failure modes.}
The blind rounds showed that the two judges err in different directions. Early on, Claude tended to align pairs that were only loosely related, while Gemini missed licensing in both directions. After the licensing rules were tightened, Gemini's dominant error became reporting a conflict when two queries differ because their questions differ, whereas Claude's more conservative reading fit the rules. Neither judge was uniformly better. Over the rounds we also compared ways of combining them, including using one judge alone, weighted averages, and falling back to one judge when the two disagree, and settled on the abstention rule of Eq.~\ref{eq:consensus_policy}: a pair on which one judge sees a conflict and the other does not is exactly where one of them is likely wrong, and we drop it from training.

\paragraph{Outcome.}
Disagreement with the adjudicator was frequent in the early rounds and fell as the rules were refined; in the last rounds the adjudicator found almost nothing to correct, and on real pairs of entries, which the judges are mainly used to label, the judge system's judgments were essentially aligned with the human's. The adjudicated pairs also exposed defects in several pair generators, which were fixed before the training data was built.

\subsection{Can Flagged Pairs Be Removed Without Review?}
To test whether the scorer's conflict score could be used to remove entries without adjudication, we audited all $252$ pairs in \texttt{california\_schools} with $\hat{\beta} \ge 0.03$ in two stages.
\begin{itemize}[leftmargin=*, itemsep=2pt]
    \item \textbf{Stage 1 (is some decision realized differently?):} the judge system, as in the first step of Track~B (§\ref{sec:two_track}), checks whether at least one matched pair of units is realized differently.
    \item \textbf{Stage 2 (do the two questions pose the same request?):} for pairs that pass Stage 1, the Track~B adjudicator checks whether the two questions ask for the same thing, in which case the difference is a bug or a convention split, or are different questions that share a template, in which case both entries are kept.
\end{itemize}

\begin{table}[h]
\centering
\caption{Two-stage audit of the flagged pairs in \texttt{california\_schools} ($n = 252$), by scorer conflict score. Stage~1 rates exclude the $36$ pairs whose judge call failed; those pairs are kept, as in Table~\ref{tab:trackb_outcomes}.}
\label{tab:calibration_tiers}
\footnotesize
\begin{tabular}{lcccc}
\toprule
\textbf{Conflict score $\hat{\beta}$} & \textbf{Pairs} & \begin{tabular}[b]{@{}c@{}}\textbf{Stage 1: some decision}\\\textbf{realized differently}\end{tabular} & \begin{tabular}[b]{@{}c@{}}\textbf{Stage 2:}\\\textbf{same request}\end{tabular} & \begin{tabular}[b]{@{}c@{}}\textbf{Different questions}\\\textbf{(shared template)}\end{tabular} \\
\midrule
$0.03 \le \hat{\beta} < 0.05$ & $149$ & $28.3\%$ ($36/127$) & $0.7\%$ ($1/149$) & $99.3\%$ \\
$0.05 \le \hat{\beta} < 0.08$ & $60$ & $45.3\%$ ($24/53$) & $3.3\%$ ($2/60$) & $96.7\%$ \\
$0.08 \le \hat{\beta} < 0.10$ & $14$ & $50.0\%$ ($6/12$) & $7.1\%$ ($1/14$) & $92.9\%$ \\
$0.10 \le \hat{\beta} < 0.15$ & $25$ & $60.0\%$ ($12/20$) & $0.0\%$ ($0/25$) & $100.0\%$ \\
$\hat{\beta} \ge 0.15$ & $4$ & $75.0\%$ ($3/4$) & $0.0\%$ ($0/4$) & $100.0\%$ \\
\midrule
All & $252$ & $37.5\%$ ($81/216$) & $1.6\%$ ($4/252$) & $98.4\%$ \\
\bottomrule
\end{tabular}
\end{table}

Stage~1 confirmation rises monotonically with the score, from $28.3\%$ to $75.0\%$, so the scorer ranks tension correctly. Stage~2 shows why flagged entries cannot be removed without review: $98.4\%$ of the flagged pairs are different questions that share a structural template (for example, two county and grade queries with different aggregations). This is why the cleaning protocol (§\ref{sec:two_track}) pairs a per-entry audit (Track~A) with rulebook adjudication that can keep both entries (Track~B).

\subsection{Worked Examples from the Judge System}
Below are three verbatim traces showing how the judge system decomposes a pair into licensed semantic units, finds conflicts, and applies the abstention rule (Eq.~\ref{eq:consensus_policy}).

\paragraph{Sample 1: a dropped predicate is novel, not a conflict (\texttt{asm\_droppred}).}
When a constructed variant drops a predicate from $S_B$ but keeps the full question $Q_B$, a mechanical diff of the two syntax trees labels the missing unit a conflict ($\beta_{\text{mech}} = 0.33$). Both judges instead compare only the units that both queries realize and place the missing unit in the novel bucket ($\omega = 0.33$, $\beta = 0.00$): the missing filter is a faithfulness failure of $e_B$ alone, the kind of defect that per-entry review (Track~A, §\ref{sec:two_track}) is meant to catch.
\begin{quote}
\small
\textbf{Pair ID:} \texttt{asm\_droppred:bird\_train:1383} \\
\textbf{$Q_A$ \& $Q_B$ (identical):} \emph{What South Asian nations have low incomes? Please include the entire names of the nations in your answer. South Asian nations refers to region = 'South Asia'; have low incomes refers to incomegroup = 'Low income'; the entire names refers to longname.} \\
\textbf{$S_A$:} \texttt{SELECT LongName FROM Country WHERE IncomeGroup = 'Low income' AND Region = 'South Asia'} \\
\textbf{$S_B$:} \texttt{SELECT LongName FROM Country WHERE Region = 'South Asia'} \\
\textbf{Scalars:} Mechanical $= (0.67, 0.33, 0.00)$ $\mid$ Claude-Sonnet-5 $= (0.67, 0.00, 0.33)$ $\mid$ Gemini-3.5-Flash $= (0.67, 0.00, 0.33)$. \\
\textbf{Alignment trace:} Both judges align $a_0 \leftrightarrow b_0$ (\emph{``South Asian nations''} $\to$ \texttt{Region = 'South Asia'}) and $a_1 \leftrightarrow b_1$ (\emph{``entire names''} $\to$ \texttt{LongName}), leaving $a_2$ (\emph{``have low incomes''} $\to$ \texttt{IncomeGroup = 'Low income'}) unmatched in $S_B$ ($\omega = 1/3$).
\end{quote}

\paragraph{Sample 2: both judges find the same conflict (\texttt{asm\_opflip}).}
When a comparison operator is flipped ($\texttt{>} \to \texttt{>=}$) under the same question, both judges independently isolate the conflicting unit ($\beta = 0.50$).
\begin{quote}
\small
\textbf{Pair ID:} \texttt{asm\_opflip:bird\_train:1733} \\
\textbf{$Q_A$ \& $Q_B$:} \emph{How many of the work orders didn't meet the due date? workers who did not meet the due date refers to EndDate > DueDate;} \\
\textbf{$S_A$:} \texttt{SELECT COUNT(WorkOrderID) FROM WorkOrder WHERE EndDate > DueDate} \\
\textbf{$S_B$:} \texttt{SELECT COUNT(WorkOrderID) FROM WorkOrder WHERE EndDate >= DueDate} \\
\textbf{Scalars:} Mechanical $= (0.50, 0.50, 0.00)$ $\mid$ Claude-Sonnet-5 $= (0.50, 0.50, 0.00)$ $\mid$ Gemini-3.5-Flash $= (0.50, 0.50, 0.00)$. \\
\textbf{Conflict rationale:} Unit $0$ (\texttt{COUNT(WorkOrderID)}) is aligned ($\alpha = 0.50$). Unit $1$ realizes the same requirement, \emph{``didn't meet the due date''}, with incompatible operators (\texttt{EndDate > DueDate} vs.\ \texttt{EndDate >= DueDate}), a conflict ($\beta = 0.50$).
\end{quote}

\paragraph{Sample 3: the judges diverge and the pair is dropped (\texttt{cross\_anchor\_real}).}
When two different questions share a join path (\texttt{club} $\bowtie$ \texttt{member\_of\_club} $\bowtie$ \texttt{student}) but filter on different attributes (\texttt{position = 'CTO'} vs.\ \texttt{major = '600'}) and select different entities, the judges can diverge on whether the structural joins should be grouped with the verbalized filters.
\begin{quote}
\small
\textbf{Pair ID:} \texttt{cross\_anchor\_real:spider:454} \\
\textbf{$Q_A$:} \emph{Show the first name and last name of the student who is the CTO of club ``Hopkins Student Enterprises''.} \\
\textbf{$S_A$:} \texttt{SELECT T3.fname, T3.lname FROM club T1 JOIN member\_of\_club T2 ON T1.clubid = T2.clubid JOIN student T3 ON T2.stuid = T3.stuid WHERE T1.clubname = 'Hopkins Student Enterprises' AND T2.position = 'CTO'} \\
\textbf{$Q_B$:} \emph{Find the club which has the largest number of members majoring in ``600''.} \\
\textbf{$S_B$:} \texttt{SELECT T1.clubname FROM club T1 JOIN member\_of\_club T2 ON T1.clubid = T2.clubid JOIN student T3 ON T2.stuid = T3.stuid WHERE T3.major = '600' GROUP BY T1.clubname ORDER BY COUNT(*) DESC LIMIT 1} \\
\textbf{Scalars:} Claude-Sonnet-5 $= (\alpha=0.38, \beta=0.25, \omega=0.38)$ $\mid$ Gemini-3.5-Flash $= (\alpha=0.18, \beta=0.00, \omega=0.82)$. \\
\textbf{Abstention:} $|\alpha_1 - \alpha_2| = 0.20$ does not exceed the alignment limit, but one judge sees a conflict and the other does not ($\beta_1 = 0.25 \ge 0.15$, $\beta_2 = 0.00 \le 0.05$), so under Eq.~\ref{eq:consensus_policy} the pair is abstained and excluded from scorer training.
\end{quote}

\section{Appendix: Cleaning Protocol Details}
\label{app:wash}

\paragraph{The rulebook.}
Table~\ref{tab:rulebook} lists the ten rules used by Track B (§\ref{sec:two_track}). The adjudicator receives each rule in full: its condition, the realization it prefers, a one-sentence reason, its exceptions, and a worked example with a winning and a losing SQL. For instance, R1 prefers \texttt{COUNT(DISTINCT X)} over plain \texttt{COUNT} when a question asks ``how many X'' and the SQL joins through a one-to-many relationship, unless the question asks for occurrences (``how many transactions''); R9, the rule cited most often, prefers the SQL that follows a definition stated in the question itself. R8 differs from the others: it marks an entry whose SQL returns no rows or only \texttt{NULL}s as low-confidence rather than naming a winner. The pairs that an earlier run of Track B settled by a rule supply the in-context experiment (§\ref{sec:p3_experiment}): after removing pairs whose question fixes the choice in words, $84$ remain, of which $53$ were settled by R1, $16$ by R6, and $15$ by other rules.

\begin{table}[h]
\centering
\caption{The Track B rulebook. Each rule applies only when its condition holds and the two SQL queries differ in the stated way.}
\label{tab:rulebook}
\footnotesize
\setlength{\tabcolsep}{4pt}
\begin{tabular}{@{}p{0.04\textwidth}p{0.39\textwidth}p{0.29\textwidth}p{0.20\textwidth}@{}}
\toprule
& \textbf{Applies when} & \textbf{Prefers} & \textbf{Exception} \\
\midrule
R1 & The question asks ``how many X'' and the SQL joins through a one-to-many relationship that can count X twice & \texttt{COUNT(DISTINCT} key of X\texttt{)} & The question asks for occurrences (``how many transactions'') \\
R2 & A ``how many X'' or ``which X'' question joins three or more tables, and one path chains sibling tables that each reference X & The join path that passes through X's own table & -- \\
R3 & The question names a specific value (place, name, code) and the two SQL use different literals & The literal that matches the question text & -- \\
R4 & A concept has both a primary column (flag or category) and a derived text column, and the two SQL use different ones & The primary column, unless the question names the derived value & -- \\
R5 & The question asks for a date range and the two SQL compare dates differently & \texttt{BETWEEN} on typed dates over string extraction & -- \\
R6 & The question asks for ``the highest/most'' (one answer) and the SQL differ in \texttt{ORDER BY \ldots\ LIMIT 1} vs.\ \texttt{= MAX(\ldots)} & \texttt{ORDER BY \ldots\ LIMIT 1} & The question asks for all tied rows \\
R7 & Arithmetic or comparison is applied to numbers stored as text, and the SQL differ in whether they cast & Casting before comparing & -- \\
R8 & A query runs but returns no rows or only \texttt{NULL}s & (Marks the entry as low-confidence; does not pick a winner) & The question asks whether any match exists \\
R9 & The question itself states a definition or mapping, and the SQL differ in whether they follow it & The SQL that follows the stated definition & -- \\
R10 & A \texttt{GROUP BY} query selects columns that are neither grouped nor aggregated & The SQL that groups or aggregates every selected column & -- \\
\bottomrule
\end{tabular}
\end{table}

\paragraph{Track A checklist.}
The Track A auditors' checklist covers defects that recur in BIRD: a percentage compared against a fraction, a strict threshold written as non-strict, a missing or unrequested filter, the wrong column for a named concept, double counting through a one-to-many join, and an impossible result, such as no rows where the question implies matches.

\paragraph{Where Track B pairs end.}
Table~\ref{tab:trackb_outcomes} shows where each candidate pair stops. On BIRD-dev, Track B keeps both entries in $1{,}043$ of the $2{,}127$ pairs at the first step and in $529$ more at the same-request check; $68$ pairs end with an internal error, $38$ with a rule, and $447$ without a decision. Of the $157$ BIRD-dev removals, $97$ come from Track A and $78$ from Track B ($18$ from both); on BIRD-train, $1{,}252$ come from Track A and $254$ from Track B. In $16$--$21\%$ of pairs a referee call failed to return a valid answer (a malformed or truncated reply after retries). These pairs count as ``no decision'' and both entries are kept, so failures can only make the protocol remove less, never more. On BIRD-train, candidates come from scoring each entry against its $20$ nearest neighbors in the same database rather than against every entry, so a conflict between two entries that are not near neighbors is not seen by Track B there.

\begin{table}[h]
\centering
\caption{Where Track B candidate pairs end. ``Keep both'' outcomes remove nothing; ``internal error'' and ``rule'' remove one side of the pair. Removed entries are counted after deduplication, since one entry can lose several pairs.}
\label{tab:trackb_outcomes}
\footnotesize
\begin{tabular}{lrr}
\toprule
& \textbf{BIRD-dev} & \textbf{BIRD-train} \\
\midrule
Candidate pairs ($\hat{\beta} \ge 0.03$) & $2{,}127$ & $3{,}101$ \\
\quad Keep both: no conflicting unit in the ACN profile & $1{,}043$ & $1{,}368$ \\
\quad Keep both: the two questions ask for different things & $529$ & $542$ \\
\quad Keep both: no error or rule found on recheck & $2$ & $2$ \\
\quad Remove one side: internal error & $68$ & $619$ \\
\quad Remove one side: rule & $38$ & $64$ \\
\quad No decision (referee call failed), keep both & $447$ & $506$ \\
\midrule
Entries removed by Track B & $78$ & $254$ \\
Entries removed by Track A (auditor consensus: bug) & $97$ & $1{,}252$ \\
Entries removed in total (either track) & $157$ & $1{,}332$ \\
\bottomrule
\end{tabular}
\end{table}

\paragraph{Fine-tuning results by difficulty.}
In Table~\ref{tab:sft_wash_ablation}, the gap between the cleaned and bad-kept sets reaches $4.0$ points at Epoch 3 and grows with difficulty, to $4.7$ on moderate and $5.2$ on challenging questions; against the random set the same ordering holds ($+1.8$ and $+3.0$). Both uncleaned sets peak at Epoch 2 or 3 and then slip, while the cleaned model holds its accuracy.

\paragraph{Training sets of the fine-tuning experiment.}
The random set of §\ref{sec:hygiene_impact} draws $5{,}648$ entries from the uncleaned pool, stratified by database and SQL length, so it keeps defective entries at their natural rate ($1{,}076$ of the $1{,}332$); it differs from the cleaned set only in which entries were dropped. The bad-kept set ($5{,}650$ entries) keeps every entry the protocol would remove and instead deletes an equal number of clean ones, matched on database and SQL length, raising the defective share from $19\%$ to $24\%$ as a dose-response check. The rise in BHI from the random to the cleaned set ($86.7 \to 90.7$) is not an artifact of removing what the scorer itself flagged: Track~A never consults the scorer, yet removing only the entries it flags raises the pool's BHI from $86.8$ to $90.3$, whereas removing as many entries at random gives $85.9$--$86.9$ over five draws.

\paragraph{Agreement with BIRD-Platinum.}
Of our removals among the $1{,}821$ questions shared with the BIRD-Platinum audit, $92.4\%$ had their SQL corrected by the experts and $97.7\%$ had some correction. The protocol removes far fewer entries than the experts corrected (it catches $33$--$38\%$ of their corrections), because it acts only on a concrete defect, a stated error, or a rule, whereas the audit also rewrote question wording and external knowledge.

\paragraph{Comparison with deleting every flagged entry.}
A simpler alternative to the protocol is to delete, without review, every entry that conflicts strongly with one peer ($\hat{\beta} \ge 0.10$) or weakly with at least two ($\hat{\beta} \ge 0.05$). On BIRD-dev this removes $152$ entries, and for $94\%$ of them the flagged partner turns out to answer a different question, as in the \texttt{california\_schools} audit (Appendix~\ref{app:calibration}). It also misses most real defects: about three quarters of the entries our protocol removes are not among the deleted ones ($117$ of $157$ on BIRD-dev, $970$ of $1{,}332$ on BIRD-train), and the two removal sets overlap by only $15$--$19\%$ (Jaccard). Fine-tuning on a BIRD-train pool cleaned this way, in an earlier run with the same design as §\ref{sec:hygiene_impact}, gave no gain over its control ($-0.1$ points).

\paragraph{Rules cited.}
On BIRD-dev the $38$ rule decisions cite R9 ($16$), R1 ($9$), R6 ($8$), R3 ($2$), R7 ($2$), and R8 ($1$); on BIRD-train the $64$ cite R1 ($23$), R9 ($18$), R6 ($13$), R3 ($3$), R5 ($3$), R4 ($2$), and R7 ($2$). A real case for R1 from BIRD-dev: ``How many users obtained the Announcer badge?'' is written as \texttt{COUNT(id)} over the badge table, while ``How many users were awarded the Citizen Patrol badge?'' is written as \texttt{COUNT(DISTINCT UserId)}. Both ask for a number of users, a user can hold a badge more than once, and R1 removes the first entry. A case for R9: two BIRD-dev entries both define abnormal total protein as \texttt{TP < 6.0 OR TP > 8.5} in their question text; one SQL follows this, the other writes \texttt{TP <= 6.0 OR TP >= 8.5}, and R9 removes the second. New rules enter the rulebook only after human review; none were proposed in the runs reported here.

\section{Appendix: In-Context Experiment Details}
\label{app:p3_details}

\subsection{Construction of the Test Sets}
\paragraph{Paired questions.}
From an earlier run of Track~B we take the pairs the adjudicator settled by a rule, in which both entries are judged defensible and the rulebook picks one. Keeping pairs whose two SQL execute and return different results, and removing duplicates and chains of pairs with contradictory labels, leaves $189$ pairs. We keep the $84$ whose questions do not settle the convention in words, dropping those that say, for example, ``as a percentage'' or ``list all tied rows''.

\paragraph{Outside questions.}
The $151$ outside questions are taken from the naturally grouped questions of the replication (Appendix~\ref{app:p3_details}, below): their wording leaves the grain open, their result changes if the grain is flipped, and whether the original gold SQL counts distinct keys or raw rows decides which examples are matched and which are opposite. They use the R1 examples, $14$ entries after removing $3$ pairs whose two sides do not use opposite counting forms; the R6 pairs are too few to form a set of examples.

\subsection{Per-Convention Breakdown}
Table~\ref{tab:p3_mirror} splits the paired questions of Table~\ref{tab:p3_results} by the convention of the gold query. Both sides move in the predicted direction: matched entries beat opposite ones by $10.0$ points on convention-A questions (exact McNemar, $p = 0.11$) and by $16.7$ points on convention-B questions ($p = 0.002$). The effect is therefore not tied to either particular convention, although at $60$ questions per side only the convention-B gap is significant. On the outside questions, matched beats opposite by $44.4$ points when the gold counts distinct keys ($n = 72$, $p = 4 \times 10^{-9}$) and by $16.5$ when it counts raw rows ($n = 79$, $p = 0.001$).

\begin{table}[h]
\centering
\caption{The in-context consistency effect on the paired questions, split by the convention of the gold query ($n = 60$ per side).}
\label{tab:p3_mirror}
\footnotesize
\begin{tabular}{lcc}
\toprule
\textbf{Context in prompt} & \textbf{Convention-A gold} & \textbf{Convention-B gold} \\
\midrule
Matched convention & $\mathbf{70.0}$ & $\mathbf{63.3}$ \\
Mixed (uncleaned bank) & $66.7$ & $55.0$ \\
None (zero-shot) & $68.3$ & $48.3$ \\
Opposite convention & $60.0$ & $46.7$ \\
\midrule
Matched $-$ opposite & $+10.0$ ($p = 0.11$) & $+16.7$ ($p = 0.002$) \\
\bottomrule
\end{tabular}
\end{table}

\subsection{Convention Families and Replication Protocol}
\paragraph{Convention families in the replication.}
The replication does not use Track~B. It groups naturally occurring BIRD queries into three convention families by their SQL form:
\begin{enumerate}[leftmargin=*, itemsep=1pt]
    \item \textbf{Entity vs.\ occurrence counting} (\texttt{COUNT(DISTINCT id)} vs.\ \texttt{COUNT(id)}, the choice R1 governs): when joining one-to-many tables (e.g., \texttt{accounts} $\bowtie$ \texttt{trans}), whether ``how many accounts'' counts distinct account IDs or matching transaction rows.
    \item \textbf{Extremum tie-breaking} (\texttt{ORDER BY x DESC LIMIT 1} vs.\ \texttt{WHERE x = (SELECT MAX(x)...)}, the choice R6 governs): whether a superlative query returns one row or all tied rows.
    \item \textbf{Ratio scaling} (\texttt{0.42} vs.\ \texttt{42.0}): whether a rate is reported as a fraction or multiplied by $100$. No rule in the rulebook covers this choice.
\end{enumerate}

\paragraph{Replication protocol.}
To complement the $84$ Track~B pairs, we classified $470$ naturally occurring BIRD queries from several databases into the three families above with deterministic pattern matchers over the SQL syntax tree, and split them by their wording into \emph{convention-open} questions, whose wording does not say whether to count distinct items, how to break ties, or whether to multiply by $100$, and \emph{convention-fixed} questions, whose wording says so explicitly (``distinct'', ``percentage'', ``top 1 row''). With Gemini-3.8-Flash, consistent examples beat mixed ones by $12.1$ points on convention-open questions ($p = 3 \times 10^{-8}$; $26$ questions flip from wrong to right and none the other way), but by only $0.8$ points on convention-fixed questions ($p = 0.82$). This supports the view that consistent examples work by settling decisions the question leaves open.

\paragraph{Prompt size.}
We ran the paired experiment in four rounds that varied the number of examples in the prompt and the number of databases they come from ($2$, $4$, $8$, about $14$ from mixed sources, and one from each of about $17$ databases). Table~\ref{tab:p3_results} reports the last configuration. With fewer examples the gaps were smaller but in the same direction, and a mix of conventions stayed close to no examples in every round. In the reported round, each exam question's own pair is excluded from its examples; earlier rounds excluded every exam pair.

\subsection{A Synthesized Bank with a Controlled Share of Conflicts}
\label{app:badbank}
This bank is deliberately synthesized to vary one property, the share of conflicting examples, while every entry stays defensible. Its twins are bank entries only, never test questions or gold SQL.

\paragraph{Questions.}
The judge system assigned a rule and a side to $4{,}597$ BIRD-train, BIRD-dev, and Spider-train entries. Both judges agree on the rule for $1{,}804$ of them; $1{,}511$ of these lie in the cleaned pools and have a question that does not fix the side, and $485$ of those return a different result when their SQL is flipped to the other side. Within each database and rule that has at least two such questions, we split them at random into seeds and tests: $206$ seeds and $246$ tests. The tests keep their original gold SQL: R1 has $64$ tests on side~A and $62$ on side~B, R6 has $113$ on A and $7$ on B, and $181$ come from BIRD and $65$ from Spider.

\paragraph{Twins.}
For each seed, Claude-Sonnet-5 writes a twin that asks for the same thing in its own words, uses the same noun for what is counted or ranked, does not state which side to use, and follows the other side of the rule with the rest of the logic unchanged. A twin is kept if its SQL executes to a result different from the seed's; if both judges, blind to the seed, assign it the seed's rule, the other side, and a question that does not fix the side; and if both judges call its SQL a reasonable answer to its question. Of the $132$ twins that pass, $75$ have scorer conflict $\hat{\beta} \ge 0.05$ with their seed ($26$ for R1 and $49$ for R6; $28$ of them from Spider, mostly R6). Each seed and its twin form a \emph{family}.

\paragraph{Banks and doses.}
For each rule there are two worlds. In world~A the base bank holds the side-A entry of every family (the original if it is on side~A, otherwise the twin), and the tests are those whose gold is on side~A; world~B mirrors this. Each dose adds the other-side entry for the first $25$, $50$, $75$, or $100\%$ of families in a fixed random order, so higher doses contain lower ones and nothing is removed. The bank grows from $75$ to $93$, $112$, $132$, and $150$ entries, of which $0$, $18$, $37$, $57$, and $75$ conflict with the test gold's convention, giving the column headers of Table~\ref{tab:badbank}. Neither side is always the synthetic one: in world~A, $28$ entries of the base bank are twins, and at the highest dose $47$ of the $75$ added entries are twins and $28$ originals, so the dose measures conflict rather than synthetic origin.

\paragraph{Prompting and BHI.}
Each test question receives $K = 8$ entries from its world's bank for its rule, first from the same database and then from others, each ranked by word overlap as in CAT; the no-context arm gives the schema only. Decoding is greedy. The share of retrieved entries that follow the gold's convention falls from $100\%$ to $85$, $70$, $60$, and $53\%$ across the doses. BHI is computed over all same-database pairs inside each bank and averaged over the two worlds, weighted by pair count.

\begin{table}[h]
\centering
\caption{Execution accuracy (\%) on $246$ held-out real BIRD and Spider questions, with their original gold SQL, as conflicting examples are added to a deliberately synthesized bank in which every entry is judged defensible. The test questions are the same in every column. The $0\%$ bank follows the same counting and superlative conventions as the test gold; each column adds more entries that follow the opposite convention (bank size $75$ to $150$), removing none. Rows retrieve the $8$ most similar bank entries; \emph{No context} gives the schema only. BHI is computed over this small, targeted bank, so its values are not comparable with Table~\ref{tab:bhi_results}; the consistent bank scores below $100$ because of its R6 entries (Table~\ref{tab:badbank_split}).}
\label{tab:badbank}
\footnotesize
\setlength{\tabcolsep}{5pt}
\begin{tabular}{lcccccc}
\toprule
\textbf{Conflicting examples (\% of bank)} & \textbf{No context} & $0$ & $19.4$ & $33.0$ & $43.2$ & $50.0$ \\
\midrule
Bank BHI & -- & $75.7$ & $20.4$ & $13.3$ & $8.5$ & $8.9$ \\
Gemini-3.8-Flash & $57.3$ & $\mathbf{75.2}$ & $68.3$ & $65.0$ & $59.3$ & $58.1$ \\
Claude-Sonnet-5 & $49.2$ & $\mathbf{70.7}$ & $66.3$ & $61.8$ & $58.9$ & $54.1$ \\
\bottomrule
\end{tabular}
\end{table}

\begin{figure}[h]
\centering
\begin{tikzpicture}
\begin{axis}[width=0.62\linewidth, height=4.6cm, xlabel={Conflicting examples in the bank (\%)}, ylabel={Execution accuracy (\%)}, xmin=-3, xmax=53, ymin=45, ymax=80, xtick={0,19.4,33.0,43.2,50.0}, xticklabel style={font=\scriptsize}, yticklabel style={font=\scriptsize}, label style={font=\small}, legend style={font=\scriptsize, at={(0.98,0.98)}, anchor=north east}, legend cell align=left]
\addplot[blue, mark=*, thick] coordinates {(0,75.2) (19.4,68.3) (33.0,65.0) (43.2,59.3) (50.0,58.1)};
\addlegendentry{Gemini-3.8-Flash}
\addplot[red, mark=square*, thick] coordinates {(0,70.7) (19.4,66.3) (33.0,61.8) (43.2,58.9) (50.0,54.1)};
\addlegendentry{Claude-Sonnet-5}
\addplot[blue, dashed, domain=-3:53, forget plot] {57.3};
\addplot[red, dashed, domain=-3:53, forget plot] {49.2};
\end{axis}
\end{tikzpicture}
\caption{Accuracy on the same 246 test questions as the share of conflicting examples in the bank grows (Table~\ref{tab:badbank}). Solid lines retrieve the 8 most similar bank entries; dashed lines give the schema only.}
\label{fig:badbank_dose}
\end{figure}
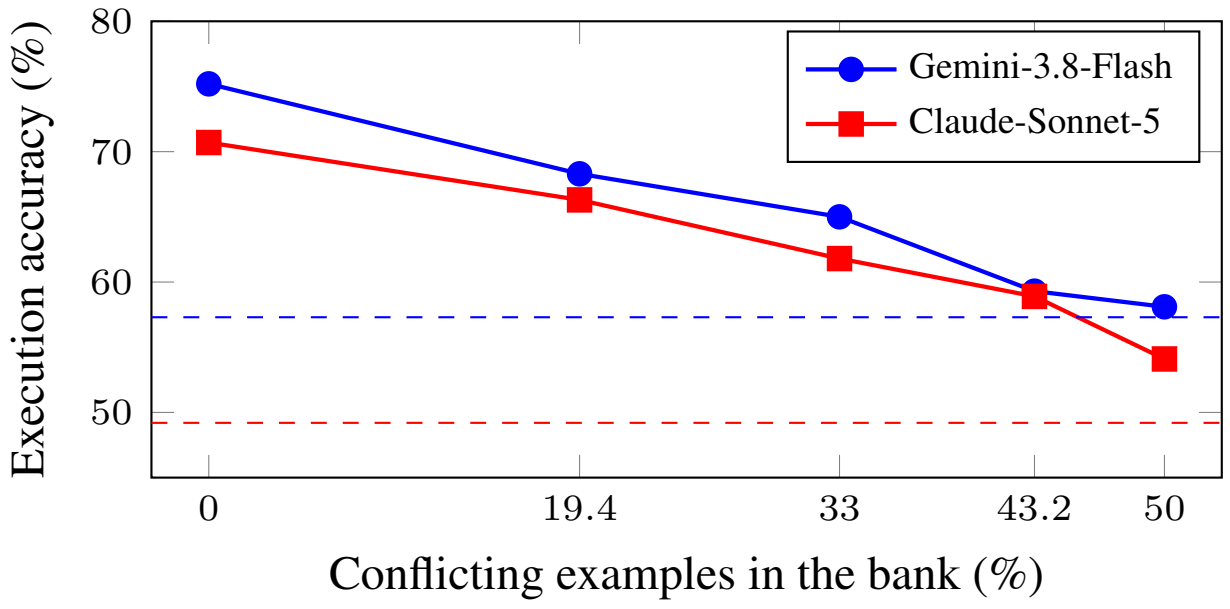

\paragraph{Results.}
Figure~\ref{fig:badbank_dose} plots Table~\ref{tab:badbank}, and Table~\ref{tab:badbank_split} splits it by rule and by source. Between the $0\%$ and $50\%$ banks, $44$ questions go from right to wrong and $2$ the other way for Gemini (exact McNemar, $p = 3 \times 10^{-11}$), and $44$ and $3$ for Claude ($p = 2 \times 10^{-10}$); the Spearman correlation between bank BHI and accuracy over the five doses is $0.90$ for both models. The drop is largest for R1, where the half-conflicting bank falls to the no-context level, and smaller for R6. Spider tests, whose no-context accuracy is already high, move little and noisily. The twins were written by Claude-Sonnet-5, one of the two evaluated models; Gemini-3.8-Flash, which wrote none of them, shows the same decline.

\begin{table}[h]
\centering
\caption{Table~\ref{tab:badbank} split by rule and by source (execution accuracy, \%). Bank BHI is per rule.}
\label{tab:badbank_split}
\footnotesize
\setlength{\tabcolsep}{5pt}
\begin{tabular}{llrcccccc}
\toprule
\textbf{Model} & \textbf{Subset} & $n$ & \textbf{No context} & $0\%$ & $19.4\%$ & $33.0\%$ & $43.2\%$ & $50.0\%$ \\
\midrule
\multirow{2}{*}{Bank BHI} & R1 & & -- & $100.0$ & $5.4$ & $1.3$ & $0.9$ & $1.2$ \\
 & R6 & & -- & $71.0$ & $26.8$ & $19.7$ & $13.5$ & $14.1$ \\
\midrule
\multirow{4}{*}{Gemini-3.8-Flash} & R1 & $126$ & $47.6$ & $73.0$ & $64.3$ & $59.5$ & $50.0$ & $48.4$ \\
 & R6 & $120$ & $67.5$ & $77.5$ & $72.5$ & $70.8$ & $69.2$ & $68.3$ \\
 & BIRD & $181$ & $49.7$ & $72.4$ & $64.1$ & $60.8$ & $51.9$ & $50.8$ \\
 & Spider & $65$ & $78.5$ & $83.1$ & $80.0$ & $76.9$ & $80.0$ & $78.5$ \\
\midrule
\multirow{4}{*}{Claude-Sonnet-5} & R1 & $126$ & $43.7$ & $69.0$ & $61.1$ & $53.2$ & $49.2$ & $43.7$ \\
 & R6 & $120$ & $55.0$ & $72.5$ & $71.7$ & $70.8$ & $69.2$ & $65.0$ \\
 & BIRD & $181$ & $45.3$ & $67.4$ & $62.4$ & $54.7$ & $50.8$ & $45.9$ \\
 & Spider & $65$ & $60.0$ & $80.0$ & $76.9$ & $81.5$ & $81.5$ & $76.9$ \\
\bottomrule
\end{tabular}
\end{table}

\section{Appendix: CAT Construction Details and CAT-Spider}
\label{app:cat_build}

\paragraph{Quality gates.}
A probe enters the benchmark only if its question contains no SQL; its gold SQL executes and returns a result that is neither empty nor a bare $0$, $1$, or \texttt{NULL}; plausible wrong queries return something different, so it cannot be answered by accident; two LLMs agree that the gold SQL answers the question; and its $\hat{\alpha}$ against its source and its $\sim 18$ nearest bank entries lies inside the group's band, with no detected conflict. For CAT-BIRD, $2{,}122$ of $3{,}074$ candidates passed, and the $500$ $d_1$--$d_3$ probes were drawn from them so that their no-context accuracy matches $d_0$, using only no-context scores, never those of a context-aware system. Distance is thus the only property that varies along the CAT-BIRD ladder by design.

\subsection{CAT-Spider}
\label{app:cat_spider}
CAT-Spider applies the same generation, gates, bands, and evaluation to Spider, with the differences below; its results should be read as a second instance of the pipeline, not as a replicate of CAT-BIRD.

\paragraph{Bank.}
The bank holds $1{,}833$ Spider-train entries of official hardness medium, hard, or extra from the $42$ databases of Table~\ref{tab:bhi_results}, after removing the entries that the Track~A auditors agree are bugs, the losing side of every pair settled by Track~B, and both sides of every pair whose adjudication failed. As in CAT-BIRD, about a quarter of each database's questions ($712$ in total), stratified by SQL length, are held out of the bank and serve only as a source of unseen logic. Unlike CAT-BIRD, every bank database comes from the training split of its benchmark.

\paragraph{Distance groups.}
Gemini-3.8-Flash and Claude-Sonnet-5 generate the probes in Spider's style. $d_0$ has $314$ rephrases from all $42$ databases ($218$ by Claude and $96$ by Gemini), each preserving its entry ($\hat{\alpha} \ge 0.70$) and conflicting with no bank entry ($\hat{\beta} < 0.05$). $d_1$ ($200$ probes, $40$ databases), $d_2$ ($149$, $41$ databases), and $d_3$ ($90$, $36$ databases) use the same bands as CAT-BIRD. $d_2$ and $d_3$ fall short of their targets of $200$ and $100$ because too few candidates that the models answer correctly without context passed the gates, so these groups could not be matched to $d_0$ on no-context accuracy and are somewhat harder without context. $d_{\max}$ holds $200$ real Spider-dev questions from its $20$ databases, which share no database with Spider-train, sampled to match $d_0$'s mix of hardness levels rather than its no-context accuracy.

\paragraph{Systems.}
Retrieval uses the same models, prompt, and top-$3$ word-overlap retrieval as CAT-BIRD. The fine-tuned system is Qwen2.5-7B-Instruct trained on the $1{,}833$ bank entries with the CAT-BIRD recipe for $20$ epochs (Appendix~\ref{app:cat_sft}). Its accuracy levels off after epoch $8$: at epochs $5$, $10$, $15$, and $20$ it scores $87.6$, $89.8$, $89.5$, and $90.1\%$ on $d_0$ and $36.7$, $33.3$, $32.2$, and $33.3\%$ on $d_3$.
The no-bank control uses the same recipe on $1{,}833$ Spider-train entries from $98$ databases outside the bank, matched to the bank on hardness. At epoch $20$ it gains $13.5$ points over the base model on $d_{\max}$, more than the bank run ($9.5$), so the bank run's $d_{\max}$ gain over the base reflects learning Spider's style rather than the bank.

\begin{table}[h]
\centering
\caption{Execution accuracy (\%) on CAT-Spider, as in Table~\ref{tab:cat_results}. Unlike CAT-BIRD, CAT-Spider's $d_2$ and $d_3$ are not matched to $d_0$ on no-context accuracy, and its $d_{\max}$ is matched on hardness levels only.}
\label{tab:cat_spider}
\footnotesize
\setlength{\tabcolsep}{5pt}
\resizebox{\textwidth}{!}{%
\begin{tabular}{llccccc}
\toprule
\textbf{Model} & \textbf{System} & $d_0$ & $d_1$ & $d_2$ & $d_3$ & $d_{\max}$ \\
\midrule
\multicolumn{2}{l}{\textbf{CAT-Spider} ($n$ per group)} & $314$ & $200$ & $149$ & $90$ & $200$ \\
\midrule
\multirow{2}{*}{Gemini-3.8-Flash} & No context & $81.2$ & $80.5$ & $73.8$ & $77.8$ & $85.5$ \\
 & Top-3 retrieved & $\mathbf{95.5}$\gain{+14.3} & $\mathbf{87.5}$\gain{+7.0} & $75.8$\gain{+2.0} & $76.7$\gain{-1.1} & $\mathbf{89.5}$\gain{+4.0} \\
\midrule
\multirow{2}{*}{Claude-Sonnet-5} & No context & $76.1$ & $75.0$ & $69.8$ & $75.6$ & $77.5$ \\
 & Top-3 retrieved & $94.6$\gain{+18.5} & $\mathbf{87.5}$\gain{+12.5} & $\mathbf{76.5}$\gain{+6.7} & $\mathbf{80.0}$\gain{+4.4} & $83.5$\gain{+6.0} \\
\midrule
\multirow{3}{*}{Qwen2.5-7B} & Base (no context) & $57.0$ & $57.0$ & $43.6$ & $41.1$ & $59.5$ \\
 & No-bank control SFT & $67.2$ & $56.0$ & $42.3$ & $38.9$ & $73.0$ \\
 & Bank SFT & $90.1$\gain{+22.9} & $70.0$\gain{+14.0} & $51.0$\gain{+8.7} & $33.3$\gain{-5.6} & $69.0$\gain{-4.0} \\
\bottomrule
\end{tabular}%
}
\end{table}

\paragraph{Uncertainty.}
Table~\ref{tab:cat_spider} gives the CAT-Spider results. $95\%$ paired bootstrap intervals for the retrieval gain are $[-3.4, 7.4]$ and $[-7.8, 5.6]$ for Gemini at $d_2$ and $d_3$, and $[1.3, 12.1]$ and $[-1.1, 11.1]$ for Claude.

\section{Appendix: Fine-Tuned Systems on CAT-BIRD}
\label{app:cat_sft}

\paragraph{Recipes.}
All runs fine-tune with LoRA (rank $128$, scaling $256$) on prompts that contain the schema and the question only, with loss on the SQL tokens, batch size $1$ with gradient accumulation $8$, and one cosine schedule over the full run. The bank runs train on the $1{,}501$ CAT-BIRD bank entries. The Qwen control trains on $1{,}501$ cleaned BIRD-train entries from $14$ databases outside CAT-BIRD, matched to the bank on per-database counts, prompt length, number of joins, and nested-query rate; its SQL is slightly longer ($177$ vs.\ $163$ characters). Qwen2.5-7B-Instruct uses peak learning rate $1\times 10^{-5}$ with $20\%$ warmup for both the bank run and the control. Gemma-4-12B-it at that rate barely moves beyond $d_0$ in $8$ epochs (Table~\ref{tab:cat_sft_epochs}); the Gemma run in Table~\ref{tab:cat_results} uses $4\times 10^{-5}$ with $10\%$ warmup. Evaluation uses greedy decoding with no retrieved context, scored by execution result-set match, after every epoch; epoch $0$ is the base model.

\paragraph{Per-epoch results.}
Table~\ref{tab:cat_sft_epochs} reports selected epochs. With $n = 200$ per group, single cells have a standard error of about $3.3$ points ($4.7$ for $d_3$), so single epochs fluctuate by $3$--$5$ points and trends are more reliable than individual cells. Averaged over epochs $10$--$20$, the Qwen bank run leads its control by $14.5$, $7.6$, and $2.9$ points at $d_1$--$d_3$. At the lower rate Gemma loses nothing on $d_{\max}$ ($69.5\%$ against $69.0\%$ at epoch $8$). Over Gemma's last ten epochs at the higher rate, its $d_{\max}$ accuracy falls from $67.5\%$ to $57.0\%$ while $d_0$ climbs from $89.7\%$ to $98.0\%$. Qwen's base model often wraps its SQL in prose, which depresses its epoch-$0$ scores; any BIRD-style fine-tuning fixes this, which is why its gains are measured against the control rather than the base: on $d_{\max}$, whose databases the bank does not cover, the bank run and the control gain $14.5$ and $12.5$ points over the base model. Qwen's training loss is about $3\times 10^{-3}$ at epoch $10$ and $10^{-4}$ from epoch $14$, so its late plateau reflects completed memorization rather than too low a learning rate. Gemma at the higher rate collapses briefly at epoch $4$, just after the learning-rate peak ($403$ of $1{,}097$ predictions fail to execute, mostly on invented or unquoted column names), and recovers at epoch $5$.

\begin{table}[h]
\centering
\caption{Per-epoch execution accuracy (\%) of fine-tuned systems on CAT-BIRD (selected epochs).}
\label{tab:cat_sft_epochs}
\footnotesize
\setlength{\tabcolsep}{5pt}
\begin{tabular}{llccccc}
\toprule
\textbf{Run} & \textbf{Epoch} & $d_0$ & $d_1$ & $d_2$ & $d_3$ & $d_{\max}$ \\
\midrule
\multirow{6}{*}{Qwen2.5-7B, bank} & 0 & $48.4$ & $35.5$ & $40.0$ & $37.0$ & $33.5$ \\
 & 2 & $65.0$ & $46.0$ & $43.5$ & $44.0$ & $47.5$ \\
 & 5 & $84.1$ & $57.5$ & $48.5$ & $46.0$ & $46.5$ \\
 & 10 & $90.9$ & $56.0$ & $44.5$ & $43.0$ & $45.5$ \\
 & 15 & $91.7$ & $55.0$ & $45.0$ & $40.0$ & $47.5$ \\
 & 20 & $92.4$ & $55.0$ & $45.5$ & $41.0$ & $48.0$ \\
\midrule
\multirow{5}{*}{Qwen2.5-7B, control} & 2 & $56.7$ & $41.0$ & $39.5$ & $45.0$ & $49.0$ \\
 & 5 & $50.4$ & $39.5$ & $38.5$ & $43.0$ & $45.5$ \\
 & 10 & $54.2$ & $36.0$ & $39.5$ & $35.0$ & $44.0$ \\
 & 15 & $55.9$ & $40.0$ & $36.0$ & $38.0$ & $44.5$ \\
 & 20 & $54.4$ & $39.5$ & $36.5$ & $40.0$ & $46.0$ \\
\midrule
\multirow{3}{*}{Gemma-4-12B, lr $1\times10^{-5}$} & 0 & $71.8$ & $67.5$ & $73.0$ & $65.0$ & $69.0$ \\
 & 4 & $80.4$ & $68.5$ & $66.0$ & $66.0$ & $66.5$ \\
 & 8 & $86.6$ & $70.5$ & $68.5$ & $66.0$ & $69.5$ \\
\midrule
\multirow{6}{*}{Gemma-4-12B, lr $4\times10^{-5}$} & 4 & $51.6$ & $34.0$ & $27.0$ & $28.0$ & $33.0$ \\
 & 5 & $77.1$ & $66.5$ & $59.5$ & $61.0$ & $64.0$ \\
 & 10 & $89.7$ & $68.0$ & $55.0$ & $61.0$ & $67.5$ \\
 & 14 & $97.2$ & $73.5$ & $61.5$ & $63.0$ & $58.5$ \\
 & 18 & $98.5$ & $74.0$ & $61.0$ & $61.0$ & $57.0$ \\
 & 20 & $98.0$ & $75.0$ & $58.0$ & $62.0$ & $57.0$ \\
\bottomrule
\end{tabular}
\end{table}

\section{Appendix: Detailed Limitations and Threats to Validity}
\label{app:limitations}

\paragraph{1. Dialect transfer and threshold portability.}
The scorer is trained mostly on SQLite: PostgreSQL contributes $5{,}133$ of its $20{,}568$ training pairs (§\ref{sec:scorer_distillation}). With an earlier scorer trained on SQLite alone, controlled probes on PostgreSQL banks (LiveSQLBench) showed the conflict channel attenuated by about $2.7\times$: convention conflicts that register $\hat{\beta} \approx 0.14$ under SQLite registered only $\approx 0.05$ under PostgreSQL. Adding PostgreSQL pairs is meant to close this gap, but we have not repeated the probe measurement for the final scorer, so the fixed thresholds of the Bank Health Index ($\hat{\beta} \ge 0.05$, $\hat{\alpha} \ge 0.50$, $1-\hat{\omega} \ge 0.1$) may still need per-dialect calibration, and the LiveSQLBench row of Table~\ref{tab:bhi_results} should be compared with the SQLite banks with this in mind. A fully dialect-balanced distillation corpus remains future work.

\paragraph{2. Benchmark construction and statistical power.}
Several features of the distance ladder bound what it can show. First, the $d_1$--$d_3$ groups were \emph{selected} so that their no-context accuracy matches $d_0$. Selection used only no-context scores and never the retrieval-augmented ones, so it cannot manufacture the retrieval gaps we report, but it does mean the flat no-context line in Table~\ref{tab:cat_results} is partly constructed rather than observed: before selection, the more distant candidate pools were somewhat harder. A reader should treat the flat baseline as a control that was imposed, not as evidence that distance and difficulty are independent in general. Second, the groups are unequal in size. The $d_3$ column rests on $100$ questions, so its cell percentages carry a Wilson $95\%$ interval of roughly $\pm 8$ points, and Gemini's $d_1 \to d_2$ step of $3$ points falls within overlapping intervals. Our headline claim is the shape of the decay across the whole ladder, which is monotone for both retrieval systems and large at its endpoints, rather than the size of any single step. Third, the distance and conflict checks compare each probe with its source and its $\sim 18$ nearest bank entries rather than all $1{,}501$, so ``no conflict'' means no detected conflict with the nearest entries. Fourth, the correctness of each generated gold query is judged by two LLMs, not verified by a human. Fifth, the probes were generated by Gemini-3.8-Flash and Claude-Sonnet-5, the same models we evaluate with retrieval, which may favor them on questions phrased in their own style; a per-generator breakdown would bound this effect. Sixth, retrieval ranks bank entries by word overlap, so at larger distances it may miss a probe's source entry, and part of the decay may reflect retriever recall rather than a limit on how far a model can carry an example. Seventh, the matched no-context difficulty need not hold for every model: the Qwen control, which never saw the bank, scores $54.4\%$ on $d_0$ but $36.5$--$40.0\%$ on $d_1$--$d_3$, which is why Qwen's gains are measured against the control. On CAT-Spider (Appendix~\ref{app:cat_spider}), the no-context difficulty is not matched at $d_2$ and $d_3$, which are also smaller ($149$ and $90$ questions), so we report it as a second instance of the decay rather than a matched replication. Finally, the fine-tuning results use one seed per run; the paired difference between the Qwen bank run and its control has a standard error of about $4.7$ points per group, so its $d_2$ gap is borderline and its $d_3$ and $d_{\max}$ gaps are within noise.

\paragraph{3. LLM referees as ground truth.}
Both cleaning tracks and the convention labels depend on LLM adjudication rather than exhaustive expert annotation. We mitigate this in three ways: consensus within the human-calibrated ensembled judge system with strict abstention (Eq.~\ref{eq:consensus_policy}) removes pairs where its judges disagree; the Track~A auditors see each entry's execution result on the live database, a non-LLM signal against queries that run but return the wrong thing; and the same-request check of Track~B (Appendix~\ref{app:calibration}) adds a second, independent stage of adjudication. Nonetheless, systematic bias shared by both judges would be invisible to a consensus protocol, and we do not claim the cleaning ledger is equivalent to expert human annotation. Two independent lines of evidence suggest the cleaning captures something real. One is annotational: $92.4\%$ of our removals were also corrected by the independent BIRD-Platinum expert audit (§\ref{sec:two_track}). The other is behavioral: the cleaned set beats a same-size random subset at every epoch of fine-tuning, and a set that keeps every defective entry does worse still, which a purely spurious labeling would not produce.

\paragraph{4. Run variance and backbone coverage.}
The SFT ablation (Table~\ref{tab:sft_wash_ablation}) reports one run per arm. We contextualize the effect against a measured evaluation noise floor of $\sigma \approx 1.3$ points: against the random control each single-epoch gap is $1.2$--$3\times$ this noise, and the effect has the same sign at all four epochs, against both controls, and in both difficulty strata; however, this is weaker than full seed-level error bars. Similarly, the in-context experiment uses a single model (Gemini-3.8-Flash) at temperature 0. The per-convention breakdown (both conventions move in the predicted direction, by $10.0$ and $16.7$ points), the independent replication on $470$ naturally grouped questions, and the test on $151$ questions outside the pairs all argue against a single-prompt artifact, but the precise magnitude of the consistency gain should be expected to vary across backbones. We therefore present the direction and rough scale of these effects as the claim, not their exact values.

\paragraph{5. What the training experiment isolates.}
The training experiment (§\ref{sec:hygiene_impact}) removes defects of both kinds, and most BIRD-train removals come from the per-entry audit (Track~A removes $1{,}252$ of the $1{,}332$). It therefore shows that cleaning helps fine-tuning, not that consistency alone does. An earlier run that deleted scorer-flagged entries without review gave no gain ($-0.1$ points; Appendix~\ref{app:wash}), which is consistent with most flags being template false alarms. Isolating consistency during fine-tuning would require a Track~B-only arm, which we leave to future work; the in-context experiment isolates it at inference time.

\end{document}